\documentclass{article} 
\usepackage{iclr2027_conference,times}

\usepackage{amsmath,amsfonts,bm}

\def\eqref#1{equation~\ref{#1}}

\def\1{\bm{1}}

\DeclareMathAlphabet{\mathsfit}{\encodingdefault}{\sfdefault}{m}{sl}
\SetMathAlphabet{\mathsfit}{bold}{\encodingdefault}{\sfdefault}{bx}{n}

\usepackage{hyperref}
\usepackage{url}
\usepackage{graphicx}
\usepackage{pifont}
\usepackage{booktabs}
\usepackage{tabularx}
\usepackage{array}
\usepackage{multirow}

\title{Brain-Conditioned Action Policies for Neural Motor Decoding}

\author{Luyao Jin$^{1}$, Running Zhao$^{2}$, Huan Zhao$^{1}$, Vincent C. K. Cheung$^{3}$, Wei-Hsin Liao$^{1}$ \thanks{whliao@cuhk.edu.hk; lyjin@link.cuhk.edu.hk} \\
$^{1}$Department of Mechanical and Automation Engineering, The Chinese University of Hong Kong\\
$^{2}$Department of Electrical and Computer Engineering, University of Hong Kong\\
$^{3}$School of Biomedical Sciences, The Chinese University of Hong Kong
}

\iclrfinalcopy 
\begin{document}

\maketitle

\begin{abstract}
Motor brain-computer interfaces (BCIs) aim to decode motor intention, enabling people with paralysis to control external devices. Neural motor decoding typically learns task-specific mappings from neural activity to kinematics, yet remains constrained by scarce paired neural–action data. We propose BrainVLA, a framework that enables neural motor decoding by drawing on a pretrained vision-language-action (VLA) model through language-mediated alignment. BrainVLA mitigates reliance on scarce paired neural–action data by leveraging VLA policies. We first construct VLA-compatible datasets including paired neural activity, action signals, language instructions, and rendered visual observations. Then, we adapt the OpenVLA-OFT policy to the target action spaces through LoRA fine-tuning. To establish an effective interface through which neural activity can convey motor intention to adapted VLA policies and guide action generation, we train a neural encoder via neural–language alignment, using language representations as semantic targets to capture latent motor intent from neural activity. The resulting neural representations serve as an endogenous intention signal to guide VLA policies to generate executable actions, while visual observations provide complementary information about the evolving task state. BrainVLA is evaluated on two neural motor datasets with different action dimensionalities using causal rollout decoding. It outperforms the evaluated baselines in cross-session decoding $R^2$ and task success rate, while demonstrating high training data efficiency. These results establish a route for neural motor decoding to draw on large-scale robotic priors through brain-conditioned VLA policies.

\end{abstract}

\section{Introduction}
Decoding motor activity from neural recordings is a central problem in brain–computer interfaces, with the potential to enable paralyzed patients to control external devices through attempted movement \cite{copilot, quadcopter}. Conventional neural decoding research has largely pursued two directions: developing increasingly capable model architectures to improve task-specific mapping from neural activity to kinematic variables \cite{NoMAD,poyo+,spint,mandt}, and scaling training across multiple neural datasets to learn more generalizable neural representations that can be adapted to downstream tasks \cite{neds,ndt3}. Nevertheless, neural motor decoding remains constrained by the limited availability of paired neural-action data. These limitations motivate us to ask: can neural motor decoding benefit from action priors within models trained on large-scale action data?

Vision-language-action (VLA) models provide a promising avenue for investigating this question, because they learn action policies from large-scale robotic demonstrations and vision-language corpora and have demonstrated strong performance across diverse robotic tasks, such as RT-2 \cite{rt-2}, OpenVLA \cite{openvla, openvla-oft}, and its variant OpenVLA-OFT \cite{openvla-oft}. However, harnessing these visuomotor policies and action priors for neural motor decoding remains underexplored. A key challenge is establishing an effective interface through which neural activity can convey motor intention to VLA policies and guide action generation. Recent studies have revealed geometric similarities between neural activity and language representations learned by language models \cite{goldstein2024,mischler2024}. These findings motivate us to explore language as a semantic bridge: aligning neural and language representations may help capture latent motor intent and enable neural activity to guide VLA policies through their original language interface. As an additional benefit of this integration, visual observations further ground the policy in the initial and evolving state.


Here, we introduce \textbf{Brain}-conditioned action policies based on the \textbf{VLA} model (\textbf{BrainVLA}), a framework that enables neural activity to condition on an adapted VLA model for motor decoding through language-mediated alignment. We first construct VLA-compatible datasets by augmenting existing paired neural-action recordings with rendered observations and generated language instructions, constructing image-language-action examples. We use these examples to adapt OpenVLA-OFT \cite{openvla-oft} with LoRA \cite{lora} to establish a language-conditioned reference policy for the target motor task. To enable neural activity to guide the policy, we train a neural encoder to align its outputs with language representations of the corresponding motor intent. This alignment provides a semantic target for learning intent-related neural representations. The adapted policy then generates actions from neural representation convey inferred motor intent, and visual observations supply complementary current task state. Together, task-specific policy adaptation and neural–language alignment allow BrainVLA to benefit neural motor decoding from pretrained robotic action priors, with the aim of improving decoding performance and training data efficiency.

We evaluate BrainVLA on two datasets with different action dimensions using causal action decoding. To assess whether decoded actions accomplish the intended tasks, we introduce task-level success criteria alongside conventional decoding $R^2$. Across both datasets, BrainVLA outperforms the evaluated baselines in cross-session decoding $R^2$ and task success rate, while demonstrating high training data efficiency. Further analyses reveal complementary modality contributions, with vision primarily supporting spatial guidance, whereas neural activity contributes more to distal motor control. In summary, the main contributions include:
\begin{itemize}
\item We introduce \textbf{BrainVLA}, a framework that decodes motor intent from neural activity, drawing on the VLA model. This provides a new strategy that leverages large-scale VLA policies for neural motor decoding.

\item We develop a language-mediated approach that aligns the neural representation with language representations to enable neural signals to guide VLA policies. This alignment provides a semantic target for capturing latent motor intention from neural activity.

\item We evaluate BrainVLA on two datasets spanning different action dimensions under a causal rollout decoding protocol. It supports zero-shot transfer to held-out sessions and remains effective in low-data regimes.
\end{itemize}

\section{Related Work}
\paragraph{Models for motor neural decoding} 
Recent motor decoding models have been developed by learning general neural representations. NDT2 \cite{ndt2} and NDT3 \cite{ndt3} use Transformer-based pretraining to capture shared structure in large-scale neural recordings, while POYO \cite{poyo} and POYO+ \cite{poyo+} employ spike-level tokenization and cross-attention to accommodate heterogeneous neural populations. NEDS jointly models neural and behavioral signals \cite{neds}, MINT leverages the geometry of population dynamics \cite{mint}, and NoMAD \cite{NoMAD}, SPINT \cite{spint}, and MANDT \cite{mandt} address cross-session variability and neural nonstationarity through latent alignment, permutation-invariant representations, or context-invariant dynamics. Collectively, these approaches primarily improve decoding by modeling, scaling, or stabilizing representations on the neural side, with behavioral variables serving mainly as decoding targets. However, they remain constrained by the limited availability of paired neural-action data. Our work mitigates reliance on scarce paired neural–action data by leveraging VLA policies.

\paragraph{Vision-language-action policies}
Vision-language-action (VLA) models condition action generation on visual observations and language instructions by integrating visual and linguistic representations within a pretrained multimodal policy. RT-2 demonstrated that knowledge acquired by large vision-language models can transfer to robotic control \cite{rt-2}, while OpenVLA provided an open-source VLA pretrained on diverse robot demonstrations and supporting parameter-efficient adaptation \cite{openvla}. Related generalist policies such as Octo emphasize efficient adaptation across embodiments \cite{octo}, while $\pi_0$ introduces flow matching for continuous, dexterous action generation \cite{pi0}. Building on OpenVLA, OpenVLA-OFT improves and stabilizes downstream adaptation \cite{openvla-oft}. Given the limitations of neural data and advances in pretrained action models, we explore VLA policies as a possibility for neural motor decoding. We therefore adopt OpenVLA-OFT as our policy backbone, leveraging its efficient and publicly reproducible fine-tuning framework to adapt a VLA policy to neural motor tasks.

\paragraph{Cross-modal neural–semantic alignment}
Cross-modal alignment provides a principled way to bridge the representational gap between neural activity and other modalities. CEBRA demonstrates that behavioral and temporal information can structure neural latent representations through contrastive learning \cite{cebra}. Recent studies further reveal substantial correspondence between neural representations during language processing and representations learned by language models, including shared embedding geometry and contextual processing hierarchies \cite{goldstein2024,mischler2024}, with model scale and context further influencing brain-language-model alignment \cite{raugel2025,gao2025}. Further, DeWave explicitly aligns EEG representations with pretrained language models for text decoding \cite{dewave}, while BIT uses contrastive alignment between neural representations and text embeddings for brain-to-text decoding \cite{bit}. In our work, we align neural representations with language representations to capture the latent motor intention from neural activity and enable neural activity to guide the VLA policy to generate executable actions.

\begin{figure}[t]
\begin{center}
\includegraphics[width=1.0\linewidth]{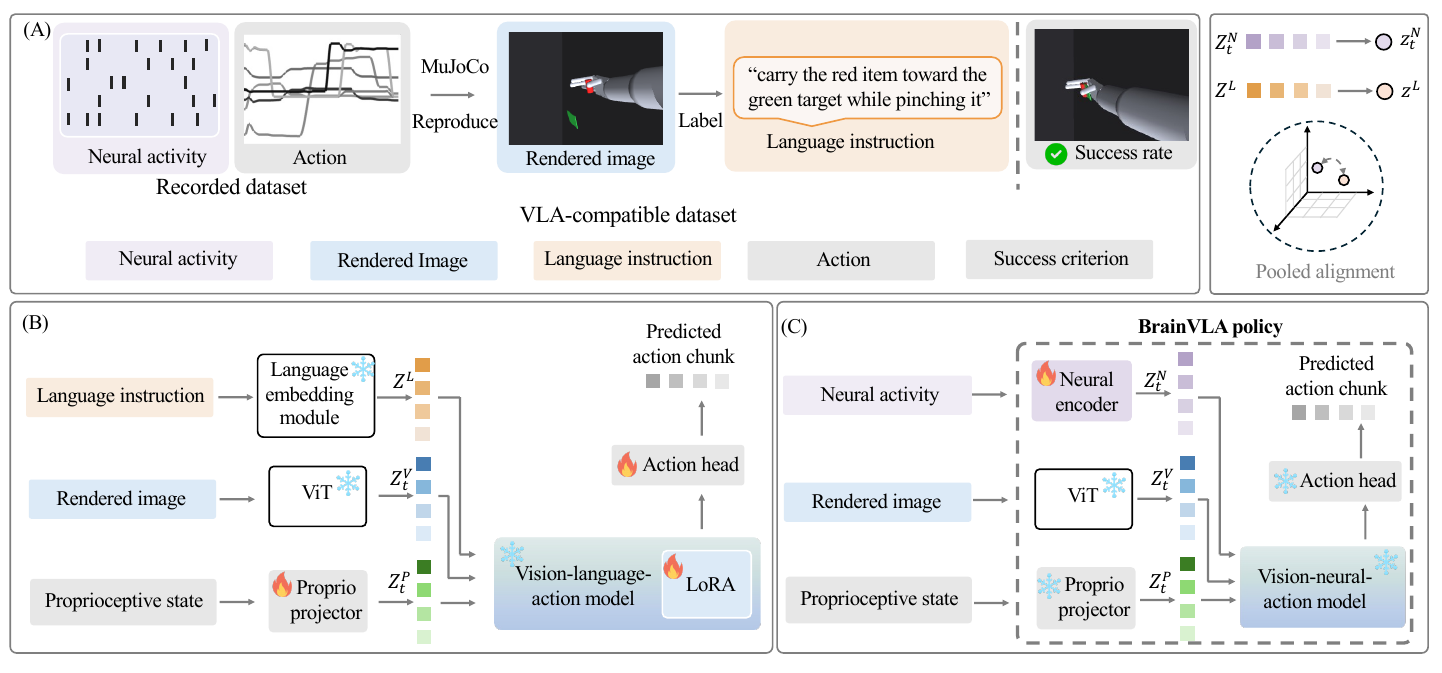}
\end{center}
\caption{Overview of BrainVLA. (A) The construction of the VLA-compatible dataset, with MuJoCo-rendered observations, language instructions, and task-specific success criteria. (B) The task-specific VLA adaptation, with language, vision, and proprioception, while task-specific modules and LoRA parameters are optimized. (C) Train a neural encoder to map neural activity into the VLA conditioning space through pooled neural-language alignment.}
\label{fig:framework}
\end{figure}

\section{Methods}
\subsection{Problem Formulation}

We consider causal rollout neural motor decoding as a mapping problem $f_{\theta}:
    \mathbf{N}_{t}
    \rightarrow
    \hat{\mathbf{A}}_t$. 
At time $t$, the decoder observes a causal neural history $\mathbf{N}_{t} = \mathbf{N}_{[t-\tau,t]}$, 
the goal is to predict an action chunk of horizon $H$: $\hat{\mathbf{A}}_t
    =
    \left[
    \hat{\mathbf{a}}_t,
    \ldots,
    \hat{\mathbf{a}}_{t+H-1}
    \right]$, using only neural information available up to the current time. Conventional neural decoders directly learn a task-specific mapping $f_{\theta}:
    \mathbf{N}
    \rightarrow
    \hat{\mathbf{A}}$.
In contrast, BrainVLA formulates neural decoding as conditioning a VLA policy on neural activity, which can utilize the action prior from the large-scale VLA policy. 
Let $g_{\phi}$ denote a task-adapted VLA policy that predicts actions from visual observations $\mathbf{o}_t$, proprioceptive state $\mathbf{p}_t$, and a conditioning representation $\mathbf{Z}^{L}$: $\hat{\mathbf{A}}_t
    =
    g_{\phi}
    \left(
    \mathbf{o}_t,
    \mathbf{p}_t,
    \mathbf{Z}^{L}
    \right)$.
During VLA adaptation, the condition signal is provided by the language embedding $\mathbf{Z}^{L}$ describing the intended action. 
BrainVLA subsequently learns a neural encoder $f_{\psi}$ that maps causal neural activity to a representation $ \mathbf{Z}^{N}_t
    =
    f_{\psi}(\mathbf{N}_{t})$
by neural-language alignment, $\mathbf{Z}^{N}_t$ and $\mathbf{Z}^{L}$ are
pooled into single-vector representations $\mathbf{z}^{N}_t$ and
$\mathbf{z}^{L}$, respectively. The alignment enables the neural representation $\mathbf{z}^{N}_t$ to capture the latent motor intention of corresponding language representations $\mathbf{Z}^{L}$. At deployment, language is no longer required; the neural representation directly replaces the language conditioning: $\hat{\mathbf{A}}_t
    =
    g_{\phi}
    \left(
    \mathbf{o}_t,
    \mathbf{p}_t,
    \mathbf{Z}^{N}_t
    \right)$.
Predicted actions are executed in MuJoCo to update the task
state and generate the subsequent visual observation $\mathbf{o}_{t+\Delta}$. Meanwhile, the causal
neural window advances from
$\mathbf{N}_{[t-\tau,t]}$ to
$\mathbf{N}_{[t+\Delta-\tau,t+\Delta]}$.
This yields a causal rollout decoding process in which neural activity provides the intended action signal, while evolving visual and proprioceptive
observations provide the current state.

\subsection{VLA-Compatible Dataset Construction}
\label{VLA-Compatible Dataset Construction}

We evaluate BrainVLA on two neural motor datasets with distinct embodiments and action spaces: Dataset 1, containing human neural activity paired with 7-DoF robotic-arm actions \cite{h1data2013,h1data2015,h1data2021}, and Dataset 2, containing non-human primate neural activity paired with 2-DoF finger-control actions \cite{m2_neuron}. Both datasets provide synchronized neural and behavioral signals, but lack the visual and language modalities required by VLA policies. To bridge the mismatch between existing neural datasets and VLA training requirements, we construct the VLA-compatible multimodal dataset. As shown in Fig. \ref{fig:framework}(A), we replay the recorded action signals in MuJoCo and render the corresponding task observations from a fixed third-person RGB camera. Each trial is additionally assigned a language instruction $\ell$ describing the intended motor behavior represented by the reconstructed vision observations. The resulting RLDS dataset contains synchronized tuples $\left(
    \mathbf{N}_t,\,
    \mathbf{o}_t,\,
    \ell,\,
    \mathbf{a}_t
    \right),$
where $\mathbf{N}_t$ denotes neural activity, $\mathbf{o}_t$ the rendered observation, $\ell$ the language instruction, and $\mathbf{a}_t$ the action target. We additionally define dataset-specific task-success criteria to complement conventional trajectory metrics $R^2$. These criteria provide a behavior-level measure of whether the decoded actions accomplish the intended motor task. These measures quantify success under kinematic reconstruction based on predicted action. For D1, success is determined using task-dependent geometric and hand-configuration criteria in MuJoCo. For D2, success is evaluated from endpoint error after integrating the predicted finger velocities. The detailed definitions are provided in Sec.~\ref{criteria}.

\subsection{Framework structure}
\paragraph{Adapted VLA policy.}
The robotic demonstrations and vision-language corpora used to pretrain the backbone OpenVLA-OFT policy are different from our neural motor tasks. We first adapt the OpenVLA-OFT policy~\cite{openvla-oft} to each neural motor task using the constructed VLA-compatible data, as illustrated in Fig.~\ref{fig:framework}(B). Given a rendered observation $\mathbf{o}_t$, language instruction $\ell$, and proprioceptive state $\mathbf{p}_t$, the frozen visual and language modules produce the corresponding visual $\mathbf{Z}^{V}_t$ and language representations $\mathbf{Z}^{L}$, while a learnable proprioception projector maps $\mathbf{p}_t$ into the policy embedding space $\mathbf{Z}^{P}_t$. These representations are then processed by the pretrained OpenVLA-OFT backbone. During adaptation, the pretrained backbone remains frozen, while LoRA adapters~\cite{lora}, the task-specific action head, and the proprioception projector are optimized because of the diverse action dimensions. This parameter-efficient adaptation accommodates the distinct state and action spaces of the two motor tasks without updating the full VLA backbone. The resulting policy predicts an action chunk $\hat{\mathbf{A}}_t
    =
    g_{\phi}
    \left(
    \mathbf{o}_t,\,
    \ell,\,
    \mathbf{p}_t
    \right)$,
where $\phi$ denotes the learnable LoRA parameters, action head, and proprioception projector. This stage establishes a task-adapted visuomotor policy that is subsequently conditioned on neural representations in BrainVLA.

\paragraph{Neural encoder.}
Following POYO~\cite{poyo}, we represent neural activity as an asynchronous sequence of spike events, where each spike token encodes its event time and unit identity. This formulation accommodates recordings with variable numbers of spikes and neural units within a shared architecture. A set of learned latent queries $\mathbf{Q}$ first attends to the spike sequence through cross-attention, compressing the variable-length neural input into a fixed-size latent representation. The resulting latent tokens are then refined through multiple self-attention blocks to capture interactions across the neural population and temporal context. Formally,
we use the final latent representation $\mathbf{Z}^{N}_t$ to represent the causal neural window $\mathbf{N}_{[t-\tau,t]}$. This representation subsequently serves as the neural conditioning input to the downstream VLA policy.

\paragraph{Neural-language alignment.}
To learn the intent-related neural representation $\mathbf{Z}^{N}_t$ and enable it to guide VLA policies, we train the neural encoder by aligning its representation with language representations of the corresponding motor intent, as illustrated in Fig.~\ref{fig:framework}(C). Because language representations usually contain different numbers of tokens, we perform alignment in a pooled embedding space. Specifically, the neural representation $\mathbf{Z}^{N}_t$ is mean-pooled over latent tokens to obtain $\mathbf{z}^{N}_t$, while the language token representation $\mathbf{Z}^{L}$ is obtained from the instruction using the frozen OpenVLA tokenizer and token-embedding module \cite{openvla-oft} and mean-pooled over valid instruction tokens to obtain $\mathbf{z}^{L}$. Both pooled representations are $\ell_2$-normalized before alignment \cite{bit}. We optimize a symmetric multi-positive contrastive objective over each minibatch \cite{contrastive_learning}. The cosine similarity between neural example $i$ and language example $j$ is $S_{ij}
    =   \left(\hat{\mathbf{z}}^{N}_{i}\right)^{\top}
    \hat{\mathbf{z}}^{L}_{j}$,
where $\hat{\mathbf{z}}^{N}$ and $\hat{\mathbf{z}}^{L}$ denote the normalized pooled embeddings. Samples sharing the same instruction are treated as positives, preventing neural examples associated with the same intended behavior from being incorrectly treated as negatives. The alignment objective is computed symmetrically in both neural-to-language and language-to-neural directions:
\begin{equation}
    \mathcal{L}_{\mathrm{align}}
    =
    \frac{1}{2}
    \left(
        \mathcal{L}_{N\rightarrow L}
        +
        \mathcal{L}_{L\rightarrow N}
    \right).
\end{equation}
The detailed parameters are in Sec. \ref{Neural–Language Alignment}. This objective encourages neural representation capture the latent motor intention of the corresponding language instruction from neural activity.

\begin{figure}[t]
\begin{center}
\includegraphics[width=1.0\linewidth]{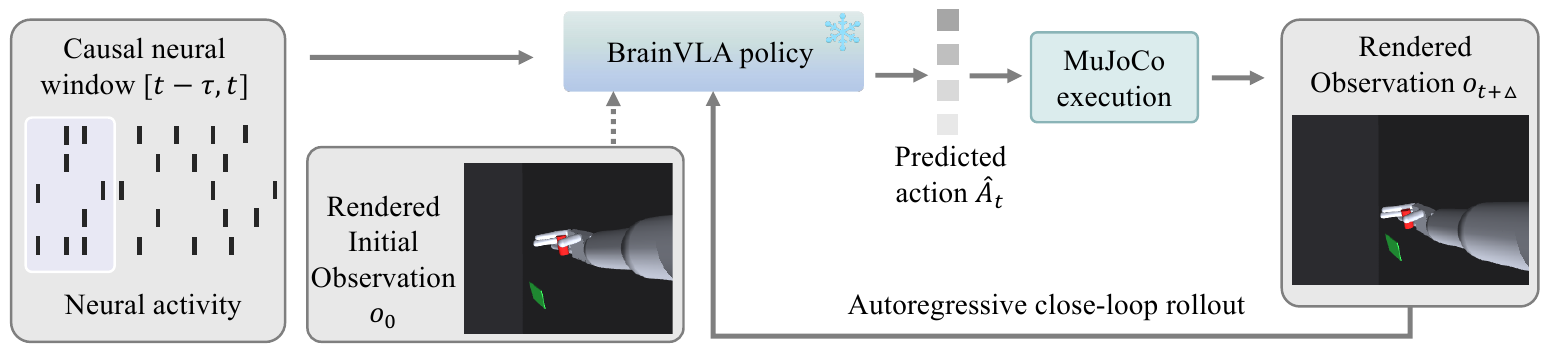}
\end{center}
\caption{BrainVLA causal rollout protocol. At each step, a causal neural window and the current MuJoCo-rendered observation are used to predict action $\hat{A}_t$. Executing $\hat{A}_t$ updates the environment and produces the next rendered observation $o_{t+\Delta}$, which is fed back to BrainVLA while the neural window advances. Thus, inference uses only past neural activity and self-evolving visual observations generated by the model's own actions.}
\label{fig:closed_loop_rollout}
\end{figure}

\paragraph{Brain-conditioned action policy.}
After neural-language alignment, the neural representation $\mathbf{Z}^{N}_t$ replaces the language conditioning $\mathbf{Z}^{L}$ of the task-adapted VLA policy. 
At time $t$, the causal neural window $\mathbf{N}_{[t-\tau,t]}$ is encoded into a neural conditioning representation $\mathbf{Z}^{N}_t$, which is provided to the policy together with the current visual observation $\mathbf{o}_t$ and proprioceptive state $\mathbf{p}_t$. 
The resulting brain-conditioned policy predicts an action chunk $\hat{\mathbf{A}}_t
    =
    g_{\phi}
    \left(
    \mathbf{o}_t,\,
    \mathbf{p}_t,\,
    \mathbf{Z}^{N}_t
    \right)$,
without requiring a language instruction at inference time.
The predicted actions are executed in MuJoCo to update the environment $o_{t+\Delta}$ and produce the subsequent observation, enabling BrainVLA to operate as a neural-conditioned visuomotor policy. 

\subsection{Training and Inference}

\paragraph{Training.}
BrainVLA is trained on the VLA-compatible RLDS dataset constructed in Sec.~\ref{VLA-Compatible Dataset Construction}. 
For each timestep $t$, the model receives a causal neural window
$\mathbf{N}_{[t-\tau,t]}$, the ground-truth rendered observation
$\mathbf{o}_t$ stored in the dataset, and the corresponding proprioceptive
state $\mathbf{p}_t$. 
The neural encoder maps the neural activity to a conditioning representation
$\mathbf{Z}^{N}_t$, which replaces the language representation in the
task-adapted VLA policy. 
The policy then predicts an action chunk
$\hat{\mathbf{A}}_t$ supervised by the recorded action target
$\mathbf{A}_t$. Training jointly optimizes action prediction and neural-language alignment:
$\mathcal{L}
    =
    \mathcal{L}_{\mathrm{action}}
    +
    \lambda_{\mathrm{align}}
    \mathcal{L}_{\mathrm{align}}$, 
with $\lambda_{\mathrm{align}}=0.01$ makes two losses are comparable.
Starting from the task-adapted VLA, we keep the VLA backbone frozen while optimizing the neural encoder. Note that training uses observations corresponding to the recorded ground-truth trajectories each cycle rather than states evolved by the model's own predictions.

\paragraph{Causal rollout inference.}
At inference time, language instruction supervision is removed, and the aligned neural representation directly conditions the VLA policy, as shown in Fig.~\ref{fig:closed_loop_rollout}. At the initial step, the neural history is zero-padded and paired with the initial observation $o_0$. The causal neural window $\tau$ is 1-s. At each subsequent query step $t$, BrainVLA receives only the causal neural history
$\mathbf{N}_{[t-\tau,t]}$, the current evolved observation
$\mathbf{o}_t$, and the current proprioceptive state $\mathbf{p}_t$:
\begin{equation}
    \hat{\mathbf{A}}_t
    =
    g_{\phi}
    \left(
        \mathbf{o}_t,\,
        \mathbf{p}_t,\,
        f_{\psi}(\mathbf{N}_{[t-\tau,t]})
    \right).
\end{equation}
The predicted action chunk is executed in MuJoCo, which updates the
environment state and produces the next visual observation
$\mathbf{o}_{t+\Delta}$ and proprioceptive state
$\mathbf{p}_{t+\Delta}$. 
The neural window then advances by the same query interval $\mathbf{N}_{[t-\tau,t]}
    \rightarrow
    \mathbf{N}_{[t+\Delta-\tau,t+\Delta]}$,
and the process repeats until the episode terminates. We use an 8-step action chunk and query every 8 steps. Unlike training, causal rollout inference does not use the ground-truth images for each query step. Instead, visual and proprioceptive inputs evolve according to BrainVLA's previously predicted actions, allowing prediction errors to affect subsequent states. This autoregressive protocol evaluates BrainVLA as a causal brain-conditioned action policy rather than as an offline neural-to-action regressor.

\section{Results}
\subsection{Experiment setup}
We split the official held-in data into approximately 80\% for training and 20\% for validation, while reserving the official held-out sessions exclusively for final testing; detailed splits are provided in Section~\ref{Training, Validation and Test splits}. BrainVLA and all reproduced baselines use the same data splits and causal rollout evaluation protocol (Fig.~\ref{fig:closed_loop_rollout}). We measure BrainVLA policy inference latency on an NVIDIA RTX PRO 6000 Blackwell Server Edition. For an 8-action prediction chunk, BrainVLA requires $0.064$\,s per query on D1 and $0.045$\,s per query on D2, respectively. Performance is evaluated using action $R^2$ and task success rate (SR) across held-out sessions. 

\begin{table}[t!]
\caption{Decoding performance comparison. The best results are highlighted in bold.}
\label{tab:decoding_performance}
\begin{center}
\begin{tabular}{l|llll}
\toprule
\multirow{2}{*}{\textbf{Method}} 
&\multicolumn{2}{c}{\bf D1} 
& \multicolumn{2}{c}{\bf D2} \\  [-3pt]
\cmidrule(lr){2-3} 
\cmidrule(lr){4-5} 
& $\bf{R^2}$ & \bf SR & $\bf{R^2}$ & \bf SR \\ [-3pt]
\midrule
Wiener filter (\cite{machine2020})&0.15$\pm$0.11&0.13$\pm$0.08&0.14$\pm$0.11&0.19$\pm$0.06 \\
RNN (\cite{machine2020})&0.32$\pm$0.16&0.21$\pm$0.17&0.18$\pm$0.21&0.26$\pm$0.12 \\
CycleGAN+WF (\cite{cycleGAN})&-0.09$\pm$0.15&0.07$\pm$0.05&-0.09$\pm$0.10&0.12$\pm$0.03 \\
POYO (\cite{poyo})&0.14$\pm$0.22&0.34$\pm$0.14&0.12$\pm$0.18&0.24$\pm$0.09 \\
NDT2 (\cite{ndt2})&0.21$\pm$0.18&0.16$\pm$0.13&0.23$\pm$0.14&0.24$\pm$0.12 \\
NoMAD (\cite{NoMAD})&0.18$\pm$0.21&0.30$\pm$0.10&0.23$\pm$0.16&0.36$\pm$0.05 \\
SPINT (\cite{spint})&0.19$\pm$0.12&0.06$\pm$0.06&0.05$\pm$0.09&0.15$\pm$0.06 \\
NDT3 (\cite{ndt3})&0.25$\pm$0.11&0.18$\pm$0.08&0.29$\pm$0.10&0.27$\pm$0.09 \\
\textbf{BrainVLA(Ours)}&\textbf{0.65$\pm$0.11}&\textbf{0.77$\pm$0.10}&\textbf{0.46$\pm$0.05}&\textbf{1.00$\pm$0.00} \\
\bottomrule
\end{tabular}
\end{center}
\end{table}

\subsection{BrainVLA Decoding performance}
\begin{figure}[t]
\begin{center}
\includegraphics[width=1.0\linewidth]{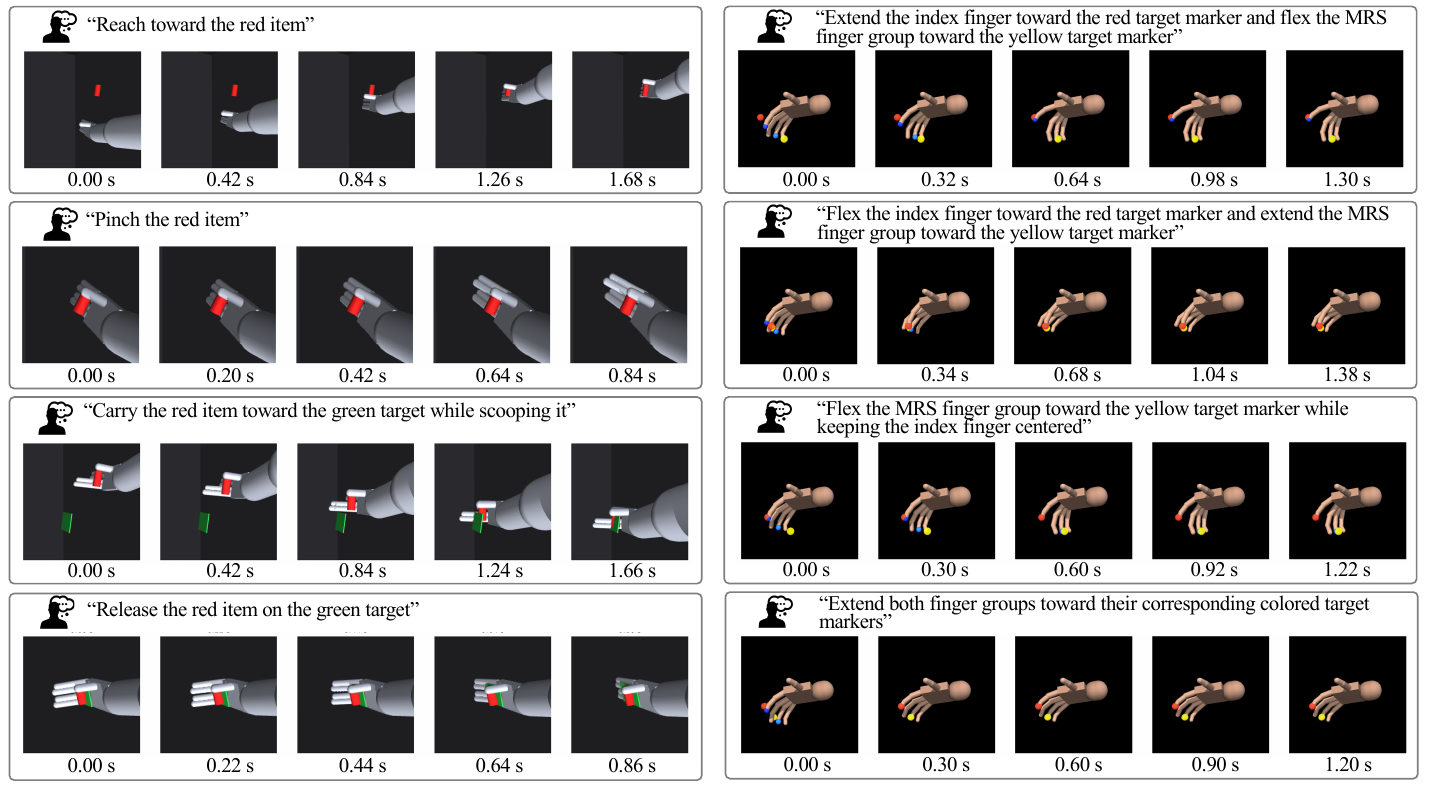}
\end{center}
\caption{Representative causal BrainVLA rollouts on D1 (left panel) and D2 (right panel). Each pane shows sequential observations from one neural-conditioned episode.}
\label{fig:real_rollout}
\end{figure}

We next evaluate whether conditioning the policy on neural activity can be effective for motor decoding. Table~\ref{tab:decoding_performance} compares BrainVLA with conventional neural decoding methods on both D1 and D2. All methods are reproduced using the same train/valid/test setting as our method. Wiener and RNN\cite{machine2020} provide conventional supervised decoding baselines; CycleGAN\cite{cycleGAN} and NoMAD\cite{NoMAD} address cross-session variability through neural representation alignment; SPINT\cite{spint} models variable neural populations using permutation-invariant unit representations; and NDT2\cite{ndt2}, NDT3\cite{ndt3}, and POYO\cite{poyo} represent recent Transformer-based neural decoding approaches. Detailed implementations are provided in Appendix~\ref{sec:baseline method}. BrainVLA achieves the strongest performance of zero-shot transfer to held-out sessions among the currently evaluated methods. D2’s high SR despite moderate $R^2$ may reflect visually guided target attainment without reproducing the reference trajectory. Fig. \ref{fig:real_rollout} shows D2 reaches the target by around the 3rd frame; differences in timing and subsequent motion can lower $R^2$ while preserving task success. The consistent gains for both datasets suggest that the proposed neural conditioning VLA policy is applicable across different action spaces and embodiments. The superior results indicate that BrainVLA has gained benefits from the VLA policy, and the language-mediated strategy is effective in enabling neural activity to guide the VLA policy.

\subsection{Closed-Loop rollout result}


Figure~\ref{fig:real_rollout} shows representative causal BrainVLA rollouts on both datasets. Each pane corresponds to a different motor episode, with frames sampled sequentially from the autoregressive rollout. The text is the corresponding language instruction, which is only used to show motor intention. Each visual observation is rendered from the state produced by previous predictions, which reflect the accumulated behavior of the policy rather than independent frame-wise predictions. The examples qualitatively demonstrate that BrainVLA can complete coherent neural-conditioned behavior over multiple rollout steps.

\subsection{Training Data Efficiency}
\begin{figure}[t]
\begin{center}
\includegraphics[width=1.0\linewidth]{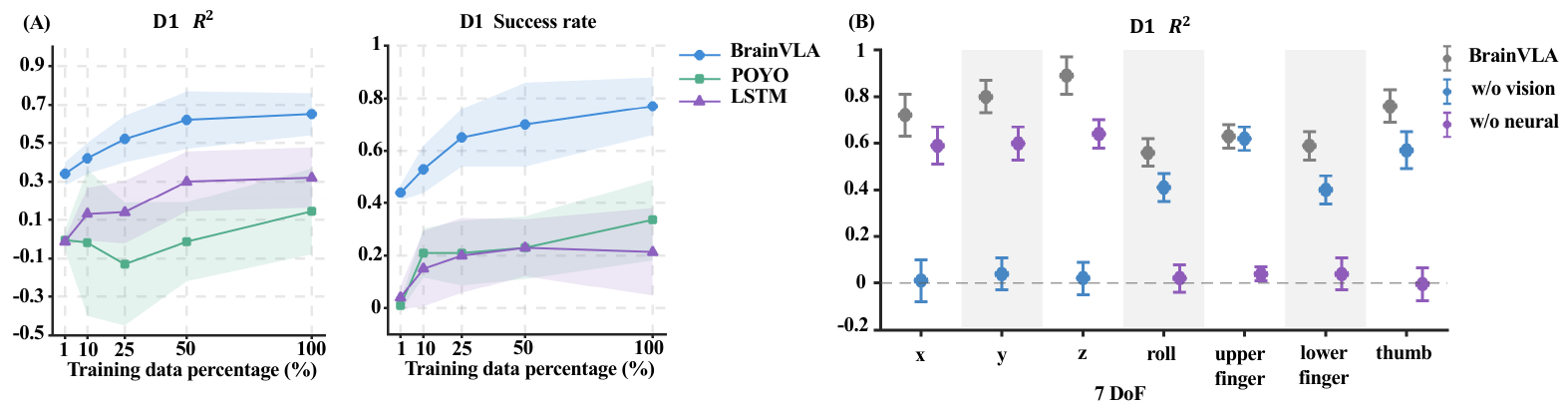}
\end{center}
\caption{(A) $R^2$ and task success rate with scaled training data. BrainVLA consistently outperforms POYO and LSTM across data regimes and remains effective under limited neural supervision. Shaded regions indicate variability across test sessions. (B) $R^2$ for each of the seven action dimensions under the full model and w/o vision or neural input. Vision removal primarily degrades the translational dimensions $(x,y,z)$, whereas neural removal more strongly affects roll and finger-related dimensions.}
\label{fig:data_efficiency}
\end{figure}

We evaluate training data efficiency by training each model with diverse fractions of the D1 training set. As shown in Fig.~\ref{fig:data_efficiency} (A), BrainVLA consistently outperforms POYO and LSTM across data regimes in both trajectory $R^2$ and task success rate. BrainVLA improves steadily with increasing training data, while the baselines exhibit substantially lower performance and greater variability, particularly in the low-data regime. Notably, BrainVLA retains meaningful decoding and task-level success even when trained with only a small subset of data. These results suggest that our policy improves the data efficiency of neural motor decoding.

\begin{table}[t]
\caption{Component ablation. Each variant is retrained and reevaluated to assess the contributions.}
\label{tab:component_ablation}
\begin{center}
\begin{tabular}{llcc|llll}
\toprule
\multirow{2}{*}{\textbf{Variant}}
& \multirow{2}{*}{\textbf{Decoder}}
& \multirow{2}{*}{\textbf{Alignment}}
& \multicolumn{1}{c|}{\multirow{2}{*}{\textbf{Vision}}} 
&\multicolumn{2}{c}{\bf D1} 
& \multicolumn{2}{c}{\bf D2} \\[-3pt]
\cmidrule(lr){5-6} 
\cmidrule(lr){7-8} 
& & & & $\bf{R^2}$ & \bf SR & $\bf{R^2}$ & \bf SR \\[-3pt]
\midrule
\bf BrainVLA&adapted VLA&$\checkmark$&Evolving&0.65&0.77&0.46&1.00 \\
w/o VLA adapt.&OpenVLA-7B&$\checkmark$&Evolving&0.39&0.43&0.34&0.75 \\
w/o language&adapted VLA&\text{\ding{55}}&Evolving&0.56&0.68&0.41&0.87\\
w/o vision&adapted VLA&\checkmark&White&0.29&0.52&0.13&0.44 \\
w/o VLA&MLP&\checkmark&Evolving&0.14&0.31&0.18&0.22 \\
\bottomrule
\end{tabular}
\end{center}
\end{table}

\subsection{Ablation studies}
We evaluate BrainVLA through two complementary ablation settings. 
First, \textbf{component ablations} isolate the contribution of major architectural and training components. Each variant is independently retrained and reevaluated under the same data splits and causal rollout protocol. These results are summarized in Table~\ref{tab:component_ablation}. 
Second, \textbf{modality ablations} examine the dependence of the trained policy on neural activity, vision, and proprioception. The BrainVLA model is kept fixed, and input modalities are removed only at inference time. The results are reported in Table~\ref{tab:modality_ablation}.


\textbf{Component Ablation.} (1) VLA adaptation. To evaluate the importance of task-specific VLA adaptation, we replace the adapted VLA backbone with the original OpenVLA-7B backbone, while keeping the neural encoder, action head, and proprioception projector trainable. The resulting degradation indicates that adapting the VLA to the target motor domain effectively establishes a language-conditioned reference policy for the brain-conditioned policy. (2) Neural-language alignment. We disable the neural-language alignment objective during training and optimize the neural encoder using only the action prediction loss. The inference pipeline remains unchanged. The performance drop shows that the alignment provides a semantic target for learning intent-related neural representations, enhancing the neural representation to guide the policy. (3) Visual observations. We remove informative vision by replacing the ground-truth rendered images with a constant white image during training and similarly replacing the evolving MuJoCo observations with the same white image during inference. This ablation therefore removes both visual state information and visual evolving feedback while leaving the remaining BrainVLA architecture unchanged. Its reduced performance highlights that visual observations supply complementary current task state, which supports the VLA policy to predict actions. (4) Adapted VLA policy. We remove the whole adapted VLA policy and replace it with a lightweight MLP decoder, while retaining its input modalities and alignment supervision. This variant excludes the adapted VLA policy as a whole, providing a direct test of large-scale VLA policy can improve action prediction. Our contribution is a framework for conditioning an adapted VLA policy on neural activity, rather than a new VLA architecture or pretraining method. The ablation experiments evaluate the effectiveness of this integration, and do not claim to isolate the effects of pretrained initialization from architectural capacity; our conclusions concern the adapted VLA policy as a whole.


\textbf{Modality Ablation.} As shown in Table~\ref{tab:modality_ablation}, the full model consistently provides the strongest performance. Removing proprioception leads to a degradation on D1, indicating that current-state information remains important even when visual feedback is available. Removing vision causes a larger loss in rollout performance, particularly on D2, highlighting the role of visual observations in tracking the evolving environment state. Similarly, removing neural activity weakens decoding performance, showing that environmental state alone is insufficient to recover the motor intention. Single-modality inference further reveals complementary roles across inputs. Neural activity alone provides intention-related information but lacks sufficient state feedback, while vision alone is insufficient to recover the full model’s decoding performance. Proprioception alone performs poorly, suggesting that instantaneous state information is insufficient without either neural intention or visual context. Overall, these results indicate that BrainVLA benefits from the complementary combination of \emph{neural intention}, \emph{visual state}, and \emph{proprioceptive state}.

\begin{table}[t]
\caption{Modality ablation. The trained BrainVLA model is kept fixed while removing modalities at inference to measure their individual and complementary contributions.}
\label{tab:modality_ablation}
\begin{center}
\begin{tabular}{lccc|llll}
\toprule
\multirow{2}{*}{\bf Variant}
&\multirow{2}{*}{\bf Neural}  
&\multirow{2}{*}{\bf Vision} 
&\multirow{2}{*}{\bf Proprio}  
&\multicolumn{2}{c}{\bf D1} 
& \multicolumn{2}{c}{\bf D2} \\ [-3pt]
\cmidrule(lr){5-6} 
\cmidrule(lr){7-8} 
& & & & $\bf{R^2}$ & \bf SR &  $\bf{R^2}$ & \bf SR \\ [-3pt]
\midrule
\bf BrainVLA&$\checkmark$&$\checkmark$&$\checkmark$&0.65&0.77&0.46&1.00 \\
Neural+Vision&$\checkmark$&$\checkmark$&\text{\ding{55}}&0.35&0.39&0.45&0.99 \\
Neural+Proprio&$\checkmark$&\text{\ding{55}}&$\checkmark$&0.27&0.29&-0.04&0.03 \\
Vision+Proprio&$\text{\ding{55}}$&$\checkmark$&$\checkmark$&0.24&0.25&0.17&0.87 \\
Neural&$\checkmark$&\text{\ding{55}}&\text{\ding{55}}&0.15&0.11&-0.06&0.43 \\
Vision&\text{\ding{55}}&$\checkmark$&\text{\ding{55}}&0.11&0.19&0.20&0.75 \\
Proprio&$\text{\ding{55}}$&$\text{\ding{55}}$&$\checkmark$&-1.01&0.03&-30.34&0.00 \\
\bottomrule
\end{tabular}
\end{center}
\end{table}

\subsection{Dimension-level analysis}
We further analyze BrainVLA on D1 at the level of individual action dimensions, as shown in Fig.~\ref{fig:data_efficiency} (B). The full model achieves consistently positive $R^2$ across all seven action dimensions, indicating that BrainVLA can recover both coarse spatial motion and more fine-grained hand configurations. The modality ablations reveal a clear complementary and functional distinction between vision and neural activity. Removing vision causes the largest degradation in the translational dimensions $(x,y,z)$, which is consistent with visual observations providing direct information about spatial configuration. In contrast, removing neural activity has a larger effect on roll and finger-related dimensions. These variables are less directly observable from the scene and depend more strongly on the intended motor command encoded in neural activity.

\section{Conclusion}
We presented BrainVLA, a framework that connects neural activity to an adapted VLA policy through neural–language alignment. By constructing VLA-compatible datasets and establishing a neural interface to the policy, this work provides an approach to using existing VLA models for neural motor decoding. The task-level success criteria are introduced to assess whether decoded actions accomplish the intended tasks to complement the decoding metric $R^2$. Across two datasets, causal rollout evaluation demonstrated strong decoding accuracy and task success on held-out sessions, alongside effective performance with limited training data. Component ablations supported the contributions of task-specific adaptation, alignment supervision, and the effectiveness of the whole adapted policy. The utilization of the VLA policy greatly improved the decoding performance, providing a promising strategy to extend neural motor decoding to a developed VLA model. Neural–language alignment not only enables the neural activity to guide the VLA policy through its language interface, but also prompts the neural encoder to extract the neural representation with latent motor intention. Modality ablations further support the complementary contributions of neural activity, vision, and proprioception to decoding performance. Dimension-level analyses further revealed complementary roles: removing vision primarily impaired translational control, whereas removing neural activity strongly affected wrist rotation and finger control. These findings demonstrate how BrainVLA realizes neural motor decoding. Finally, BrainVLA requires validation in real-time closed-loop BCI settings, where sensory feedback may alter neural activity. Our results represent an initial step toward practical neural control using adapted VLA policies.

\bibliography{iclr2027_conference}
\bibliographystyle{iclr2027_conference}

\appendix 

\newpage
\appendix
\section{Dataset and Evaluation Details}
\subsection{Training, Validation and Test splits}
\label{Training, Validation and Test splits}
We construct the data split at the session level. For each dataset, the sessions listed in Table~\ref{tab:Validation session} are selected as the validation set because their trials account for approximately 20\% of the held-in data, allowing the remaining sessions to retain roughly 80\% of the data for model training. The official held-out sessions are reserved for final testing and are not used for either model training or model selection. This protocol ensures that hyperparameter tuning and checkpoint selection are performed only on held-in data, while evaluation on the held-out sessions provides an independent assessment of generalization to unseen sessions.

\begin{table}[h]
\centering
\caption{Validation session for both datasets.}
\label{tab:Validation session}
\small
\begin{tabular}{l}
\toprule
\textbf{D1} \\
\midrule
held-in-calib\_ses-19250108T111022 \\
held-in-calib\_ses-19250119T114045 \\
held-in-calib\_ses-19250120T115537 \\
\midrule
\textbf{D2} \\
\midrule
held-in-calib\_ses-2020-10-27-Run2\_behavior+ecephys\\ 
\bottomrule
\end{tabular}
\end{table}

\subsection{Task definitions and language instructions}
\label{app:language_instructions}

We summarize the language instructions associated with the D1 and D2 datasets in Tables~\ref{tab:d1_instructions} and~\ref{tab:d2_instructions}, respectively. D1 spans seven task labels and 18 distinct instruction strings. D2 spans initialization, outward movement, and return movement, with 17 distinct suggested instruction strings. The tables preserve the source wording while using placeholders to compactly represent repeated phrases. These placeholders are notation for this appendix and are not part of the instruction strings.

\paragraph{D1: object manipulation.}
The D1 instructions describe reaching toward a red item, orienting the
wrist, shaping the hand, acquiring the item, carrying it toward a green
target, reorienting it, and releasing it.
The descriptions distinguish wrist rotation direction and acquisition
mode (pinching, grasping, or scooping).

\begin{table*}[b]
\centering
\caption{D1 language instructions.
\texttt{[direction]} is \texttt{clockwise} or
\texttt{counter-clockwise};
\texttt{[action]} is \texttt{pinch}, \texttt{grasp}, or \texttt{scoop};
and \texttt{[holding]} is \texttt{pinching}, \texttt{grasping}, or
\texttt{scooping}. All six holding-direction combinations occur
under \texttt{Orient2}.''}
\label{tab:d1_instructions}
\small
\renewcommand{\arraystretch}{1.15}
\begin{tabularx}{\textwidth}{
    @{}l>{\raggedright\arraybackslash}X@{}
}
\toprule
Task label & Instruction template \\
\midrule
Reach &
reach toward the red item \\

Orient &
rotate the wrist \texttt{[direction]} to align the hand with the red item \\

Shape &
move the thumb inward \newline
spread the thumb \\

Grasp &
\texttt{[action]} the red item \\

Carry &
carry the red item toward the green target while \texttt{[holding]} it \\

Orient2 &
keep \texttt{[holding]} the red item and rotate the wrist
\texttt{[direction]} to align with the green target \\

Release &
release the red item on the green target \\
\bottomrule
\end{tabularx}
\end{table*}

\paragraph{D2: finger-group movements.}
The D2 descriptions specify movements of the index finger, the MRS
finger group, or both groups. The index finger is associated with a
red target marker and the MRS group with a yellow target marker.
Outward instructions describe movement toward the target markers,
whereas return instructions explicitly describe movement back toward
the center target markers. Single-group instructions also specify
that the other group remains centered. For movements involving both
groups, the descriptions distinguish motion in the same direction
from simultaneous flexion and extension in opposite directions.

\begin{table*}[t]
\centering
\caption{D2 suggested language instructions.
\texttt{[motion]} is \texttt{flex} or \texttt{extend}.
For opposite-direction movements, \texttt{[opposite]} is
\texttt{extend} when \texttt{[motion]} is \texttt{flex}, and vice versa.
Each movement template represents two observed strings.}
\label{tab:d2_instructions}
\small
\renewcommand{\arraystretch}{1.15}
\begin{tabularx}{\textwidth}{
    @{}ll>{\raggedright\arraybackslash}X@{}
}
\toprule
Phase & Group / pattern & Instruction template \\
\midrule
Initialization & Both &
hold both finger groups at their corresponding colored center
target markers \\
\midrule
Outward & Index &
\texttt{[motion]} the index finger toward the red target marker
while keeping the MRS finger group centered \\

Outward & MRS &
\texttt{[motion]} the MRS finger group toward the yellow target marker
while keeping the index finger centered \\

Outward & Both / same &
\texttt{[motion]} both finger groups toward their corresponding
colored target markers \\

Outward & Both / opposite &
\texttt{[motion]} the index finger toward the red target marker
and \texttt{[opposite]} the MRS finger group toward the yellow
target marker \\
\midrule
Return & Index &
\texttt{[motion]} the index finger back toward the red center
target marker while keeping the MRS finger group centered \\

Return & MRS &
\texttt{[motion]} the MRS finger group back toward the yellow center
target marker while keeping the index finger centered \\

Return & Both / same &
\texttt{[motion]} both finger groups back toward their corresponding
colored center target markers \\

Return & Both / opposite &
\texttt{[motion]} the index finger back toward the red center
target marker and \texttt{[opposite]} the MRS finger group back
toward the yellow center target marker \\
\bottomrule
\end{tabularx}
\end{table*}

\subsection{MuJoCo environment details}
\subsubsection{MuJoCo Environment Details for D1 dataset}
We reconstruct the motor tasks in MuJoCo using a lightweight hand workspace environment \cite{h1data2015}. The simulator uses a timestep of $0.02$\,s and zero gravity. The workspace contains a planar floor and three surrounding walls, together with a task board defining source and target locations for object manipulation. The hand model is represented by a movable root with three translational degrees of freedom and a wrist roll joint. Hand articulation additionally includes a thumb joint and two finger-cluster joints controlling pinch and scoop motions. The corresponding joint ranges and scene geometry are summarized in Table~\ref{tab:mujoco_scene}.

Task objects are represented using red cylindrical source objects and green target regions. Four source-target pairs are distributed along the task board. Additional mocap bodies are used to represent the currently active pickup object and target during replay. Two fixed cameras are defined in the scene. We use the \texttt{front} camera for the main visual observations in our experiments, while the \texttt{closeup} camera is retained for visualization and debugging. Exact camera poses are listed in Table~\ref{tab:mujoco_scene}. Image preprocessing and policy-query settings are described separately in the implementation details.

Visual observations were obtained from a single fixed third-person frontal camera in MuJoCo. The camera was positioned at (1.06, 0.72, 1.00) in world coordinates, with an approximately (45$^\circ$) vertical field of view. RGB observations were rendered at (256 $\times$ 256) pixels and 50 Hz, center-cropped using 90\% of the image area, and resized to (256 $\times$ 256) before being provided to the VLA policy. The policy was queried every eight simulation steps (6.25 Hz).

\begin{table}[t]
\centering
\caption{MuJoCo scene configuration used for D1 reconstruction.}
\label{tab:mujoco_scene}
\small
\begin{tabular}{ll}
\toprule
\textbf{Parameter} & \textbf{Setting} \\
\midrule
Simulation timestep & $0.02$ s \\
Gravity & $(0,\,0,\,0)$ \\
Render size & $256\times256$ \\
\midrule
Hand root position & $(0.52,\,0.28,\,0.93)$ \\
$x$ translation range & $[-0.55,\,0.55]$ \\
$y$ translation range & $[-0.35,\,0.35]$ \\
$z$ translation range & $[-0.35,\,0.35]$ \\
Wrist roll range & $[0.35,\,3.14]$ rad \\
Thumb joint range & $[-3.14,\,3.14]$ rad \\
Pinch joint range & $[0,\,1.1]$ rad \\
Scoop joint range & $[0,\,1.1]$ rad \\
\midrule
Task-board position & $(0.0,\,0.10,\,0.95)$ \\
Task-board orientation & $(0,\,0,\,-1.6)$ rad \\
Source-object geometry & Cylinder, radius $0.028$, half-length $0.045$ \\
Target geometry & Cylinder, radius $0.045$, half-length $0.006$ \\
Number of source-target locations & $4$ \\
\midrule
Front camera position & $(1.06,\,0.72,\,1.00)$ \\
Front camera \texttt{xyaxes} &
$(0.72,\,-1.10,\,0;\;0.05,\,0.02,\,-0.99)$ \\
Closeup camera position & $(0.98,\,0.58,\,1.08)$ \\
Closeup camera \texttt{xyaxes} &
$(0.64,\,-0.77,\,0;\;0.24,\,0.16,\,-0.96)$ \\
\bottomrule
\end{tabular}
\end{table}

\begin{table}[t]
\centering
\caption{MuJoCo scene configuration used for D2 reconstruction.}
\label{tab:mujoco_m2_scene}
\small
\begin{tabular}{ll}
\toprule
\textbf{Parameter} & \textbf{Setting} \\
\midrule
Simulation timestep & $0.02$ s \\
Gravity & $(0,\,0,\,0)$ \\
Render size & $256\times256$ \\
\midrule
Controlled DoF & Index, MRS finger group \\
Index joint range & $[0,\,1.35]$ rad \\
MRS joint range & $[0,\,1.35]$ rad \\
Input representation & Normalized position increments \\
Replay mapping & Cumulative sum $\rightarrow$ clip $[0,1]$ $\rightarrow$ $[0,1.35]$ rad \\
Center configuration & $(0.675,\,0.675)$ rad \\
Extended configuration & $(0,\,0)$ rad \\
Flexed configuration & $(1.35,\,1.35)$ rad \\
\midrule
Hand position & $(0.08,\,0,\,0.035)$ \\
Hand orientation & $(-0.8236,\,0,\,-0.3491)$ rad \\
Index target geometry & Sphere, radius $0.018$ \\
MRS target geometry & Sphere, radius $0.020$ \\
\midrule
Front camera position & $(0.05,\,-0.70,\,0.32)$ \\
Top camera position & $(0.02,\,0,\,0.72)$ \\
Closeup camera position & $(-0.04,\,-0.53,\,0.25)$ \\
Camera target & Hand body \\
\midrule
Optional velocity gain & $k_v=30$ \\
Velocity control range & $[-8,\,8]$ \\
\bottomrule
\end{tabular}
\end{table}

\subsubsection{MuJoCo Environment Details for D2 dataset}
For the 2-DoF finger-control dataset, we reconstruct the task using a lightweight MuJoCo hand model with two independently controlled degrees of freedom \cite{m2_neuron}. The first joint controls the index finger, while the second controls a coordinated middle-ring-small (MRS) finger group. Both joints are modeled as hinge joints with a range of $[0,1.35]$~rad.

The recorded behavioral signals represent normalized position increments at 20 ms intervals. During replay, these increments are cumulatively integrated, clipped to $[0,1]$, and linearly mapped to the corresponding joint range $[0,1.35]$~rad. The simulator uses a timestep of $0.02$~s with zero gravity. The reconstructed hand contains a fixed thumb together with articulated index
and MRS finger groups. Two task targets are represented by mocap-controlled
spheres corresponding to the index and MRS movements. Fingertip sites and
visual markers are defined for both controlled finger groups to facilitate
target placement and visualization. The scene provides three fixed cameras---\texttt{front}, \texttt{top}, and
\texttt{closeup}---all targeting the hand. Their positions and the principal
simulation parameters are summarized in Table~\ref{tab:mujoco_m2_scene}. Visual observations were obtained from a single fixed third-person frontal camera in MuJoCo.

\subsection{Task-success evaluation criteria}
\label{criteria}

We evaluate task success using dataset-specific criteria applied to
replayed decoder predictions. D1 uses task-dependent geometric and
hand-configuration criteria in MuJoCo, whereas D2 uses endpoint error
after integrating predicted finger velocities. These measures quantify
simulated replay or kinematic success; they do not establish closed-loop
control with feedback to the neural decoder.

\paragraph{D1.}
Each task segment receives a binary success label according to
Table~\ref{tab:h1_success}.
Let $d_t^{\mathrm{item}}$ denote the Euclidean distance between the
hand-associated item reference point and the simulated item, and let
$d_t^{\mathrm{target}}$ denote the distance between the item and its
target. Distances are measured in simulation world coordinates.
Subscript $T$ denotes the final evaluated state.

The final roll error is
\[
e_T^{\mathrm{roll}}
=
\left|
\operatorname{wrap}_{[-\pi,\pi)}
\left(r_T-r_T^{\mathrm{GT}}\right)
\right|,
\]
computed from the recorded pose roll coordinate before its mapping to
the simulated joint. Thus, this criterion compares roll with the
ground-truth endpoint, rather than measuring full object-orientation
error. The closure signal $c_T$ is the final scoop coordinate when the
instruction contains ``scoop,'' the pinch coordinate when it contains
``pinch,'' and the maximum of these two coordinates otherwise
(with scoop taking precedence if both occur).

\begin{table}[ht]
\centering
\small
\caption{D1 success criteria using the evaluator's default thresholds.
All conditions within a row must hold.}
\label{tab:h1_success}
\begin{tabular}{@{}ll@{}}
\toprule
Task & Success criterion \\
\midrule
Reach
& $\min_t d_t^{\mathrm{item}} \leq 0.055$ \\
Orient
& $d_T^{\mathrm{item}} \leq 0.075$
  and $e_T^{\mathrm{roll}} \leq 0.35$ \\
Shape
& $|q_T-q_T^{\mathrm{GT}}| \leq 0.18$
  and direction agreement as defined below \\
Grasp
& $\min_t d_t^{\mathrm{item}} \leq 0.065$
  and $c_T \geq 0.60$ \\
Carry
& $d_T^{\mathrm{target}} \leq 0.070$
  and $c_T \geq 0.60$ \\
Orient2
& $d_T^{\mathrm{target}} \leq 0.070$
  and $e_T^{\mathrm{roll}} \leq 0.35$ \\
Release
& $d_T^{\mathrm{target}} \leq 0.075$
  and $c_T \leq 0.35$ \\
\bottomrule
\end{tabular}
\end{table}

For Shape, $q$ is the pinch, scoop, or thumb coordinate selected
from the instruction, in that priority order. If none is specified,
the coordinate with the largest absolute ground-truth change is used.
Writing $\Delta q=q_T-q_0$ and
$\Delta q^{\mathrm{GT}}=q_T^{\mathrm{GT}}-q_0$, direction agreement is
\[
|\Delta q^{\mathrm{GT}}| < 0.05
\quad \text{or} \quad
\operatorname{sgn}(\Delta q)
=
\operatorname{sgn}(\Delta q^{\mathrm{GT}}).
\]
Reach and Grasp use the minimum item distance over the segment;
their proximity condition need not hold at the endpoint.
These rules use geometric and configuration proxies without an
additional contact, force, or sustained-hold requirement.
Unsupported task tags are excluded from the scored set.
For supported tasks, unavailable required measurements produce failure;
measurement summaries use finite values and take the last finite value
as the final measurement.

\paragraph{D2.}
Each trial is evaluated in the two-dimensional normalized position
space of the index and middle--ring--small (MRS) finger groups.
Let $\mathbf{x}_0^{\mathrm{GT}}$ be the recorded initial position,
$\widehat{\mathbf{v}}_t$ the predicted per-bin displacement, and
$\mathbf{g}$ the target recorded in the first pose row of the trial.
For a trial containing $N$ prediction bins, positions are reconstructed as
\[
\widehat{\mathbf{x}}_0
=
\operatorname{clip}_{[0,1]^2}
\left(\mathbf{x}_0^{\mathrm{GT}}\right),
\qquad
\widehat{\mathbf{x}}_{t+1}
=
\operatorname{clip}_{[0,1]^2}
\left(\widehat{\mathbf{x}}_t+\widehat{\mathbf{v}}_t\right),
\quad t=0,\ldots,N-2.
\]
Clipping is applied independently to each coordinate. Predictions are
treated as per-bin displacements, with no additional time-step factor.
The success label is
\[
S_{\mathrm{M2}}
=
\mathbb{I}
\left[
\left\|
\widehat{\mathbf{x}}_{N-1}-\mathbf{g}
\right\|_2 < 0.15
\right].
\]
Only the final distance determines success; reaching the target earlier
does not suffice. The implementation uses $N-1$ updates for $N$
prediction bins, so the last predicted displacement is not integrated.
Trials without corresponding pose data are skipped.

\paragraph{Aggregation.}
For each dataset, the overall success rate is
\[
\mathrm{SR}
=
\frac{1}{|\mathcal{E}|}
\sum_{i\in\mathcal{E}} S_i,
\]
where $\mathcal{E}$ contains the scored D1 task segments or evaluated D2
trials. This is a pooled episode-level rate, rather than an unweighted
average of task-level or session-level rates.

\section{Implementation and Training Details}
\subsection{VLA Adaptation Details}
We adapt the OpenVLA-7B backbone to our neural motor datasets using parameter-efficient fine-tuning. The configuration for D1 is summarized in Table~\ref{tab:vla_finetuning}, and the configuration for D1 is summarized in Table \ref{tab:openvla-d2-finetune}. Three components are optimized during adaptation: the LoRA adapters inserted into the pretrained VLA backbone, the action head, and the proprioceptive projector, while the remaining backbone parameters are kept frozen. LoRA is configured with rank $r=32$, scaling parameter $\alpha=16$, and zero dropout. The model is trained for 50,000 optimization steps using AdamW with a constant learning rate of $5\times10^{-4}$ and an effective batch size of 8. The action head predicts an action chunk of length 8 in parallel using an $\ell_1$ action-regression objective. Overall, the adaptation setting is almost the same for the two datasets except for the action dimensionality, while preserving the majority of the pretrained OpenVLA backbone. Training used BF16
precision on one NVIDIA RTX PRO 6000 Blackwell Server Edition GPU.

\begin{table}[t]
\centering
\caption{OpenVLA fine-tuning configuration for D1.}
\label{tab:vla_finetuning}
\begin{tabular}{ll}
\hline
\textbf{Parameter} & \textbf{Setting} \\
\hline
Base model & OpenVLA-7B \\
Fine-tuning method & LoRA + action head + proprioceptive projector \\
Optimization steps & 50{,}000 \\
Effective batch size & 8 \\
Gradient accumulation steps & 1 \\
Optimizer & AdamW \\
Learning rate & $5 \times 10^{-4}$ \\
Learning-rate schedule & Constant over 50{,}000 steps \\
LoRA rank & 32 \\
LoRA scaling parameter $\alpha$ & 16 \\
LoRA dropout & 0.0 \\
Total trainable parameters & 278.76M \\
Training objective & L1 action regression \\
Action decoding & Parallel \\
Action chunk length & 8 \\
Action dimensionality & 7 \\
Visual input & One RGB image, $224 \times 224$ \\
Proprioceptive dimensionality & 7 \\
Action/proprioception normalization & 1st--99th percentile bounds \\
Image augmentation & Enabled \\
Numerical precision & BF16 \\
\hline
\end{tabular}
\end{table}

\begin{table}[t]
\centering
\caption{OpenVLA fine-tuning configuration for D2.}
\label{tab:openvla-d2-finetune}
\begin{tabular}{ll}
\toprule
\textbf{Parameter} & \textbf{Setting} \\
\midrule
Base model & OpenVLA-7B \\
Fine-tuning method & LoRA + action head + proprioceptive projector \\
Optimization steps & 50{,}000 \\
Effective batch size & 8 \\
Gradient accumulation steps & 1 \\
Optimizer & AdamW \\
Learning rate & $5 \times 10^{-4}$ \\
Learning-rate schedule & Constant over 50{,}000 steps \\
LoRA rank & 32 \\
LoRA scaling parameter $\alpha$ & 16 \\
LoRA dropout & 0.0 \\
Total trainable parameters & 194.79M \\
Training objective & L1 action regression \\
Action decoding & Parallel \\
Action chunk length & 8 \\
Action dimensionality & 2 \\
Visual input & One RGB image, $224 \times 224$ \\
Proprioceptive dimensionality & 2 \\
Action/proprioception normalization & 1st--99th percentile bounds \\
Image augmentation & Enabled \\
Numerical precision & BF16 \\
\bottomrule
\end{tabular}
\end{table}

\subsection{Neural Encoder}
The encoder follows a POYO-style latent-query architecture. For an action query at time $t$, the neural encoder receives spikes from
the causal lookback interval $[t-1\,\mathrm{s},\,t]$. Each spike is represented by its neural-unit identity and timestamp relative to the beginning of the lookback window. Different from common POYO, we use the same unit identity for the units across sessions, thus do not need any calibration and achieve zero-shot. Unit-start and unit-end boundary tokens are added to expose the set of recorded units, including units that do not spike in the selected window. Ordinary spikes and boundary tokens are distinguished using learned token-type embeddings. For input token $i$, the non-temporal representation is
\begin{equation}
    \mathbf{x}_i
    =
    \mathbf{E}_{\mathrm{unit}}[u_i]
    +
    \mathbf{E}_{\mathrm{type}}[c_i],
    \qquad
    \mathbf{x}_i \in \mathbb{R}^{256},
\end{equation}
where $u_i$ is the neural-unit index and $c_i$ denotes the token type.
Spike timing is incorporated through rotary time embeddings in the
attention layers.

The neural encoder configuration is summarized in Table~\ref{tab:neural-encoder-config}. Then, it uses 38 learned latent queries with timestamps uniformly distributed over the one-second encoding interval. The latent queries first cross-attend to the variable-length spike-token sequence. The resulting latent representations are then processed by four self-attention layers with eight attention heads per layer. The neural representation dimension is 256, and the final projector consists of layer normalization followed by a linear transformation:
\begin{equation}
    \mathbb{R}^{256}
    \longrightarrow
    \mathbb{R}^{4096}.
\end{equation}
The output of the bridge for a batch of size $B$ is therefore
\begin{equation}
    \mathbf{Z}
    \in
    \mathbb{R}^{B \times 38 \times 4096}.
\end{equation}

\subsection{Neural–Language Alignment}
\label{Neural–Language Alignment}
The neural-language alignment configuration is given in Table~\ref{tab:alignment_config}. During alignment, the 38 neural latent embeddings are mean-pooled to obtain a single neural representation, while the language representation is obtained by masked mean pooling over valid instruction-token embeddings from the frozen OpenVLA language embedding module. Both representations are $\ell_2$-normalized before computing cosine similarity. We optimize a symmetric multi-positive contrastive objective in both neural-to-language and language-to-neural directions, where samples associated with the same instruction are treated as positive pairs. Training uses a batch size of 32 and a learnable logit scale initialized as $1/0.07$ and capped at 100.

\subsection{Fixed-Slot Neural Prompt}
\label{sec:fixed_slot_neural_prompt}

We condition the adapted OpenVLA policy using a fixed-length neural soft prompt. For each action query at time $t$, the neural bridge maps the
preceding one second of spiking activity to
\begin{equation}
    \mathbf{Z}_t
    =
    f_{\theta}\!\left(\mathcal{S}_{t-1:t}\right)
    \in \mathbb{R}^{38 \times 4096},
\end{equation}
where each of the 38 output vectors has the same dimensionality as an
OpenVLA language-token embedding. The policy prompt comprises a 10-token textual prefix, 38 neural embedding slots, and a 5-token textual suffix:
\begin{equation}
\underbrace{
    \texttt{<s> In: What action should the robot take to}
}_{10\ \text{tokens}}
\;
\underbrace{
    [\mathbf{z}_{t,1},\ldots,\mathbf{z}_{t,38}]
}_{38\ \text{neural tokens}}
\;
\underbrace{
    \texttt{?\textbackslash n Out:}
}_{5\ \text{tokens}}
\end{equation}
Thus, the policy prompt always contains $10 + 38 + 5 = 53$ tokens. The policy additionally receives one RGB observation and
normalized proprioception. Thus, the frozen policy is conditioned jointly on three modalities:
\begin{equation}
    \text{policy input}
    =
    \{
        \text{RGB image},
        \text{proprioception},
        \text{neural soft prompt}
    \}.
\end{equation}
The neural embeddings replace only the 38 designated language-embedding
positions. Visual features are computed by OpenVLA's vision pathway,
while normalized proprioception is incorporated through its
proprioceptive projector, detailed shown in Table \ref{tab:fixed_neural_prompt}. OpenVLA, the action head, and the
proprioceptive projector remain frozen; only the neural bridge is
optimized in this experiment.

Although an auxiliary instruction string is temporarily supplied to the
standard OpenVLA preprocessing pipeline to construct an action-training
batch, the resulting textual prompt is subsequently discarded and
reconstructed using the fixed 53-token layout. This auxiliary string
therefore does not provide semantic task information to the policy.

\begin{table}[t]
  \centering
  \caption{Neural encoder configuration.}
  \label{tab:neural-encoder-config}
  \begin{tabular}{ll}
    \toprule
    Setting & Value \\
    \midrule
    Model dimension & 256 \\
    Latent queries & 38 \\
    Encoder latents & 38 (\texttt{num\_encoder\_latents}\(=\!0\)) \\
    Cross-attention & 1 head, head dim 64, RoPE on Q/K/V \\
    Self-attention depth & 4 \\
    Self-attention heads & 8, head dim 64 (inner dim 512) \\
    Dropout & 0.1 \\
    Rotary time range & \([10^{-4},\, 4.0]\), rotate dim 32 \\
    LLM projector & LayerNorm + linear \(256 \rightarrow 4096\) \\
    Output shape & \([B,\, 38,\, 4096]\) \\
    \bottomrule
  \end{tabular}
\end{table}

\begin{table}[t]
\centering
\caption{Neural-language alignment configuration.}
\label{tab:alignment_config}
\small
\begin{tabular}{ll}
\toprule
\textbf{Parameter} & \textbf{Setting} \\
\midrule
Embedding dimension $d$ & 4096 \\
Batch size $B$ & 32 \\
Neural pooling & Mean over latent slots $\mathbf{n}_i=\frac{1}{M}\sum_{m=1}^{M}\mathbf{Z}^{N}_{i,m}$ \\[3pt]
Language pooling & Masked mean over valid instruction tokens $\mathbf{t}_i=
\frac{\sum_{\ell}m_{i,\ell}\mathbf{T}_{i,\ell}}
{\sum_{\ell}m_{i,\ell}}$ \\[6pt]
Embedding normalization & $\ell_2$ normalization $\hat{\mathbf{n}}_i=\mathbf{n}_i/\|\mathbf{n}_i\|_2,\quad
 \hat{\mathbf{t}}_i=\mathbf{t}_i/\|\mathbf{t}_i\|_2$ \\[5pt]
Similarity function & Cosine similarity $S_{ij}=\hat{\mathbf{n}}_i^{\top}\hat{\mathbf{t}}_j$ \\[3pt]
Contrastive formulation & Symmetric multi-positive \\
Positive-pair definition & Same instruction ID $P_{ij}=1[y_i=y_j]$ \\[3pt]
Alignment directions & Neural$\rightarrow$Language, Language$\rightarrow$Neural $\mathcal{L}_{\mathrm{align}}
=\frac{1}{2}
(\mathcal{L}_{N\rightarrow L}
+\mathcal{L}_{L\rightarrow N})$ \\
Logit scale & Learnable, $s=\exp(\alpha)$ \\
Logit-scale initialization & $1/0.07 \approx 14.29$ \\
Maximum logit scale & 100 \\
Language embedding module & Frozen OpenVLA token embeddings \\
\bottomrule
\end{tabular}
\end{table}

\begin{table}[t]
\centering
\caption{
Composition of the fixed neural prompt supplied to OpenVLA.
The task instruction is not included in the policy input.
}
\label{tab:fixed_neural_prompt}
\small
\begin{tabularx}{\linewidth}{
    >{\raggedright\arraybackslash}p{1.7cm}
    >{\raggedright\arraybackslash}p{1.3cm}
    >{\raggedright\arraybackslash}p{1.6cm}
    X}
\toprule
Segment & Positions & Length & Content and embedding source \\
\midrule
Text prefix
& 1--10
& 10
& \texttt{<s> In: What action should the robot take to};
standard frozen OpenVLA token embeddings. \\

Neural slots
& 11--48
& 38
& One 4096-dimensional embedding per slot, produced from causal
spiking activity by the neural bridge. Placeholder lookup embeddings
are completely replaced. \\

Text suffix
& 49--53
& 5
& \texttt{?\textbackslash n Out:\textvisiblespace};
standard frozen OpenVLA token embeddings. \\
\midrule
Policy prompt
& 1--53
& $10+38+5=53$
& Fixed across tasks; no task-language tokens are supplied to the policy. \\

Action targets
& 54--109
& 56
& Eight future actions $\times$ seven action dimensions, represented
using OpenVLA action tokens during training. \\

Stop token
& 110
& 1
& End-of-sequence supervision used during action training. \\
\midrule
Training sequence
& 1--110
& 110
& 53 prompt tokens, 56 action tokens, and one stop token. \\
\bottomrule
\end{tabularx}
\end{table}

\section{Baseline Implementation Details}
\subsection{Baseline Methods}
\label{sec:baseline method}

\paragraph{Wiener filter.}
The Wiener filter \cite{machine2020} is a linear decoder that estimates behavior from the recent history of neural population activity. Spike trains are converted into binned firing rates, optionally smoothed, and concatenated across temporal lags to form an input vector. The behavioral estimate is computed as
$\hat{\mathbf{y}}_t = \mathbf{b} + \sum_{\ell=0}^{L-1}\mathbf{W}_{\ell}\mathbf{x}_{t-\ell}$,
where $\mathbf{x}_{t-\ell}$ denotes neural activity at lag $\ell$.
The parameters are fitted by minimizing squared prediction error, with ridge regularization to limit overfitting. During inference, the learned mapping directly converts the current and preceding neural observations into behavioral predictions.

\paragraph{RNN.}
Recurrent neural networks model the nonlinear temporal relationship between neural activity and behavior through a recurrent hidden state \cite{machine2020}. At each time step, the network combines the current spike-count vector with its previous state, summarizes how neural activity has evolved during the preceding second, and a readout layer converts the resulting representation into a behavioral estimate. Our implementation uses a unidirectional long short-term memory (LSTM) network, whose gates regulate the retention and updating of temporal information. The model is trained end-to-end using mean squared error between predicted and observed velocities.

\paragraph{CycleGAN.}
CycleGAN addresses cross-session recording variability by learning an unsupervised transformation between source-session and target-session neural activity. Two generators map activity in opposite directions, while two discriminators encourage the translated activity to match the corresponding session distributions. A cycle-consistency loss encourages reconstruction of the original activity after the two mappings are applied consecutively, reducing arbitrary transformations. A behavioral decoder is first trained using labeled source data and then frozen. At inference, target-session activity is mapped into the source representation and passed to the fixed decoder. This procedure requires target neural activity for alignment but does not require target behavioral labels.

\paragraph{NoMAD.}
Nonlinear Manifold Alignment with Dynamics (NoMAD) stabilizes decoding by aligning neural activity through a shared model of latent dynamics. The original method first fits a Latent Factor Analysis via Dynamical Systems (LFADS) model to reference-session recordings and trains a behavioral decoder on the inferred latent activity. For a new session, an alignment network learns to map the observed neural activity into coordinates compatible with the reference dynamics. Unsupervised objectives encourage agreement with the reference latent distribution and reconstruction of the observed spikes, while the reference dynamics and behavioral decoder remain fixed. The aligned latent trajectories are subsequently decoded using the reference behavioral mapping.

\paragraph{SPINT.}
The Spatial Permutation-Invariant Neural Transformer (SPINT) \cite{spint} treats a recorded neural population as an unordered set of units. A shared neural identity encoder extracts context-dependent unit embeddings from unlabeled calibration activity. These embeddings are combined with each unit's recent activity, and cross-attention aggregates information across units to predict behavioral variables. Dynamic channel dropout during training exposes the model to different population compositions and encourages robustness to missing units. When applied to a new session, SPINT computes new unit embeddings from unlabeled neural observations while keeping its learned parameters fixed, enabling adaptation without gradient updates or behavioral labels.

\paragraph{NDT2.}
Neural Data Transformer 2 (NDT2) \cite{ndt2} learns transferable representations of spiking activity through multi-context pretraining. Neural recordings are discretized into spike counts, and groups of channels are combined into spatial patches at each time step. The resulting tokens encode neural activity together with temporal, spatial, and recording-context information. During pretraining, a subset of tokens is masked, and the model learns to reconstruct the missing activity using a spike-count likelihood objective. The pretrained representation is subsequently adapted for downstream tasks, including behavioral decoding. This process allows information learned across recording sessions and subjects to support decoding with limited task-specific data.

\paragraph{NDT3.}
Neural Data Transformer 3 (NDT3) \cite{ndt3} extends large-scale neural modeling through autoregressive training on paired neural activity and behavior. Binned spike counts are grouped into spatial patches, while behavioral dimensions are represented as separate tokens. Modality-specific projections and position embeddings organize these tokens into a sequence processed by a causal Transformer. The model learns to predict subsequent tokens from the available preceding context, allowing neural and behavioral information to be modeled jointly. After pretraining across diverse recordings, the model is fine-tuned on the downstream task to produce behavioral predictions. Evaluation must specify whether previous behavioral inputs are observed, predicted, or unavailable, because this choice changes the decoding setting.

\paragraph{POYO.}
POYO is a neural population decoding framework that represents individual spikes as tokens associated with neuronal identity and event time \cite{poyo}. An input cross-attention module compresses the variable-length spike sequence into a fixed collection of latent representations, which are processed by latent self-attention layers. Output queries associated with the desired decoding times then attend to these representations to generate behavioral predictions. The model is trained using paired neural recordings and behavioral targets across sessions. Its tokenization and latent bottleneck support recordings with different neuronal populations, while adaptation to a new recording can involve learning new unit embeddings and fine-tuning the decoder.

\subsection{Reproduction and Causal Adaptation Details}
The reproduced experiments for all baseline methods use the same held-in training and validation sessions, and held-out calibration recordings for evaluation as our methods. The documented implementations include a ridge-regularized Wiener filter, a two-layer LSTM with 256 hidden units per layer, CycleGAN alignment followed by a frozen Wiener filter, and SPINT with context-dependent unit embeddings. The local NoMAD baseline is a causal GRU-based adaptation with maximum mean discrepancy alignment, and the local NDT2 baseline is a compact causal spatiotemporal Transformer trained from scratch; these should therefore be identified as NoMAD-inspired and NDT2-style implementations, respectively. CycleGAN and the NoMAD-inspired model use unlabeled held-out neural activity for alignment, while SPINT uses it to infer unit identities without parameter updates; these constitute transductive calibration settings. Neural models in this experiment are trained from scratch for at most 30 epochs, with early stopping after six non-improving validation evaluations. Performance is measured using variance-weighted $R^2$ in original velocity units and task-specific simulated replay or endpoint success.

\section{Additional Quantitative Results}
\subsection{Language-Conditioned Closed-Loop Rollout results}

\begin{table}[t]
\caption{VLA adaptation with language and vision (oracle policy). Task head specifies learnable
action heads and proprioception projectors.}
\label{tab:vla_adaptation}
\begin{center}
\begin{tabular}{l|llll}
\hline
\multicolumn{1}{c|}{\bf Policy} 
&\multicolumn{1}{c}{\bf D1 ($\bf{R^2}$)}
&\multicolumn{1}{c}{\bf D1 (SR)}
&\multicolumn{1}{c}{\bf D2 ($\bf{R^2}$)}
&\multicolumn{1}{c}{\bf D2 (SR)}
\\ \hline
OpenVLA-7B + task heads&0.86$\pm$0.03&0.85$\pm$0.06&0.39$\pm$0.02&0.99$\pm$0.01 \\
OpenVLA-7B + LoRA + task heads&0.89$\pm$0.02&0.97$\pm$0.04&0.49$\pm$0.02&1.00$\pm$0.00 \\
\hline
\end{tabular}
\end{center}
\end{table}

\begin{figure}[t]
\begin{center}
\includegraphics[width=1.0\linewidth]{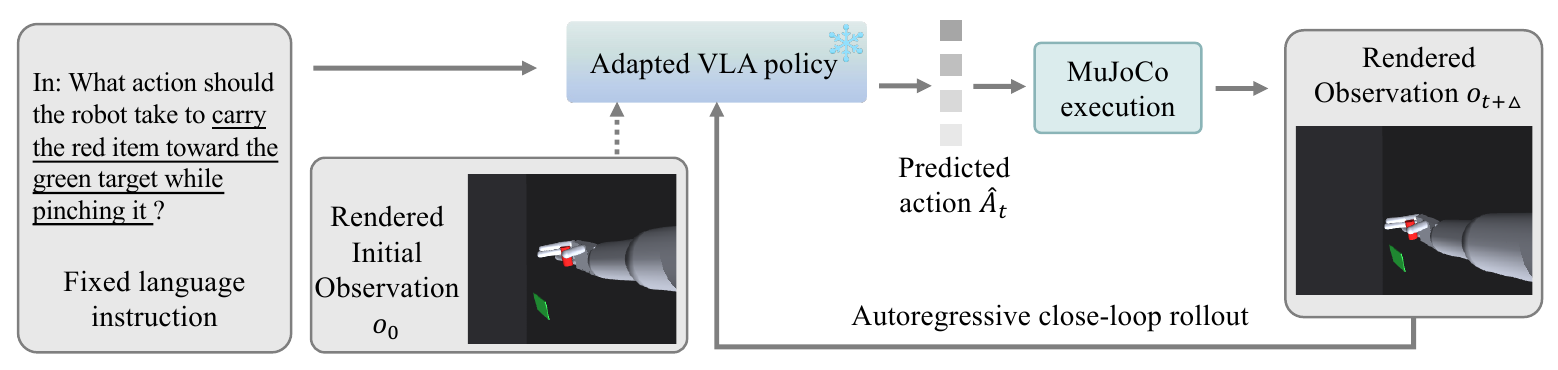}
\end{center}
\caption{Language-conditioned closed-loop rollout protocol. At each step, the fixed language instruction and the current MuJoCo-rendered observation are used to predict action $\hat{A}_t$. Executing $\hat{A}_t$ updates the environment and produces the next rendered observation $o_{t+\Delta}$, which is fed back to the VLA policy. Thus, inference uses only past fixed language instruction and self-evolving visual observations generated by the model's own actions.}
\label{appen:language_close_loop_rollout}
\end{figure}

We first evaluate whether the pretrained VLA backbone can be adapted to our motor tasks before applying neural conditioning. In this oracle setting, the policy receives the language instruction and the current visual observation, shown in Fig. \ref{appen:language_close_loop_rollout}. Because the native OpenVLA action does not directly match the action spaces of D1 and D2, both policies in Table~\ref{tab:vla_adaptation} use a frozen OpenVLA-7B backbone with a learnable task-specific action head and proprio projectors. The second additionally introduces LoRA modules to adapt the pretrained VLA
representation. The results show that task-specific adaptation produces an effective visuomotor policy for BrainVLA, while LoRA further improves its performance.

\subsection{Brain-conditioned Closed-Loop Rollout results}
\begin{figure}[h]
\begin{center}
\includegraphics[width=1.0\linewidth]{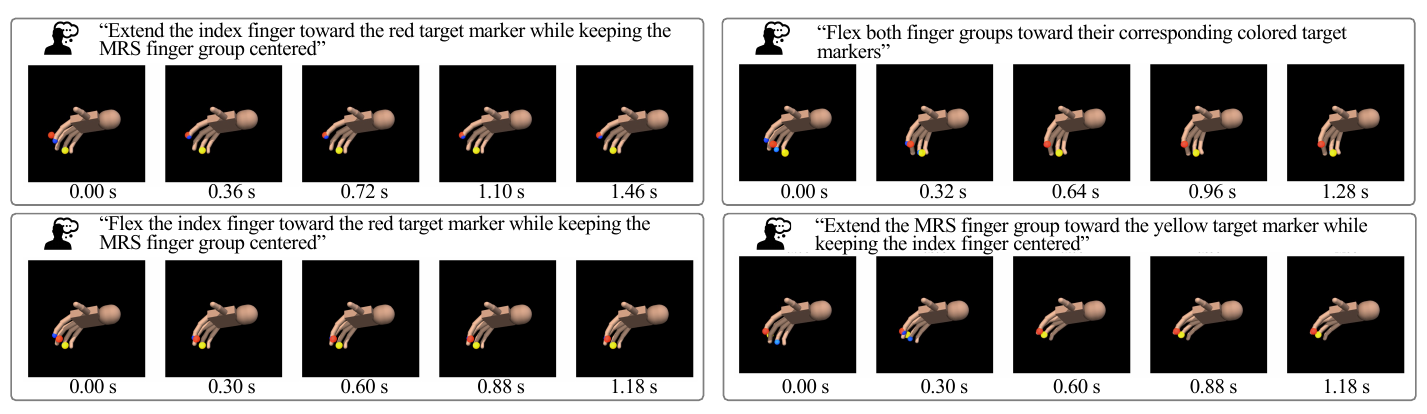}
\end{center}
\caption{Representative closed-loop BrainVLA rollouts on D2. Each row shows sequential observations from one neural-conditioned episode, progressing from left to right. At every step, BrainVLA predicts an action from the causal neural history and current rendered observation; the action is executed in MuJoCo to generate the next observation. The examples illustrate progressive reaching, object interaction, and target-directed manipulation under autoregressive visual feedback.}
\label{fig:real_rollout_D2}
\end{figure}

Due to space limitations, the main paper presents only a small number of representative closed-loop rollouts. Additional qualitative results for both D1 and D2 are provided in the appendix, as shown in Figs.~\ref{fig:real_rollout_D1} and~\ref{fig:real_rollout_D2}. These examples cover a broader range of reaching, grasping, object manipulation, wrist rotation, and finger-control behaviors. In each rollout, BrainVLA predicts an action from the causal neural history together with the current visual observation, and the predicted action is executed in MuJoCo to generate the next observation. The additional examples further illustrate the temporal evolution of neural-conditioned behavior under autoregressive visual feedback.

\begin{figure}[h]
\begin{center}
\includegraphics[width=1.0\linewidth]{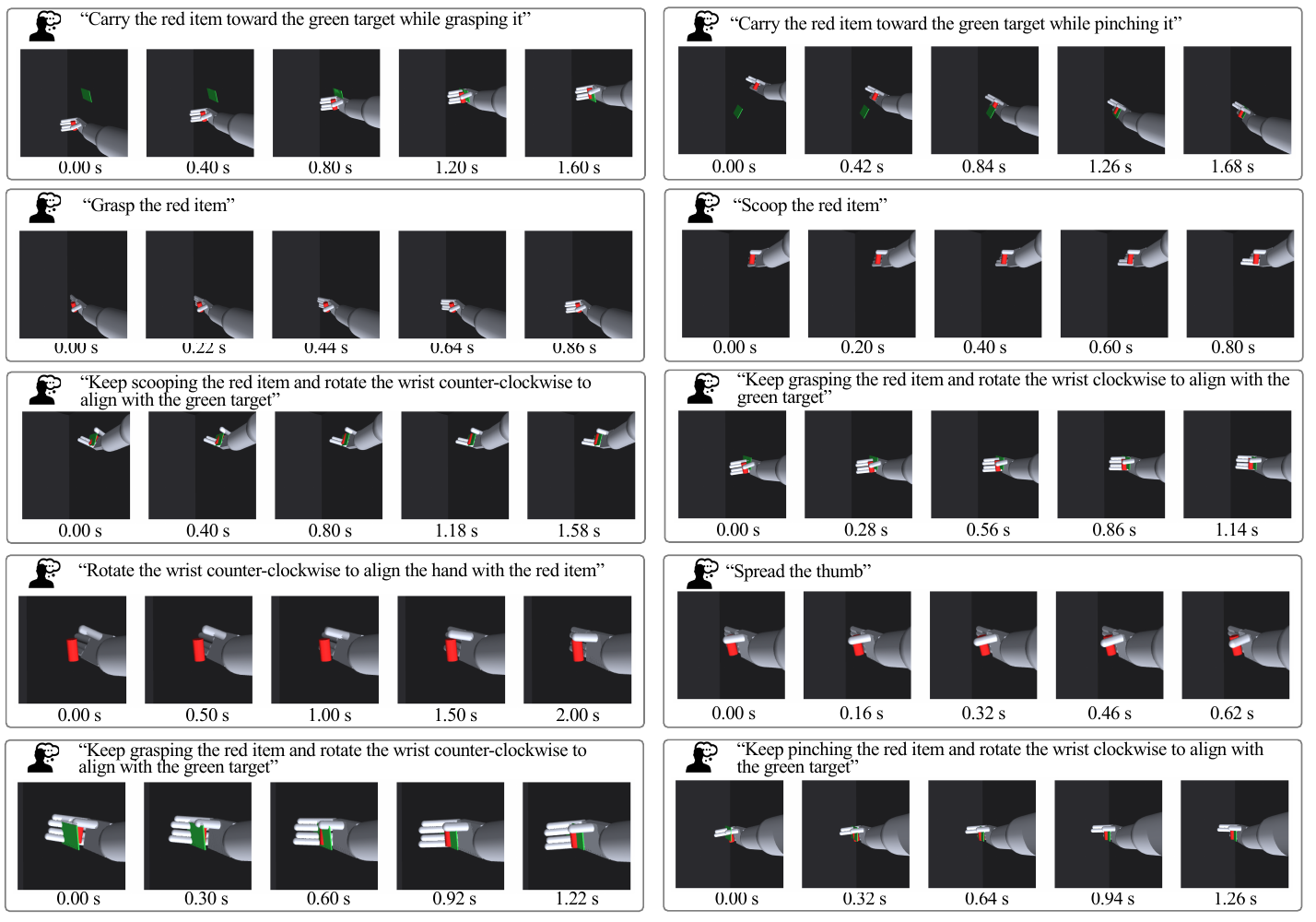}
\end{center}
\caption{Representative closed-loop BrainVLA rollouts on D1. Each row shows sequential observations from one neural-conditioned episode, progressing from left to right. At every step, BrainVLA predicts an action from the causal neural history and current rendered observation; the action is executed in MuJoCo to generate the next observation. The examples illustrate progressive reaching, object interaction, and target-directed manipulation under autoregressive visual feedback.}
\label{fig:real_rollout_D1}
\end{figure}

\begin{figure}[h]
\begin{center}
\includegraphics[width=1.0\linewidth]{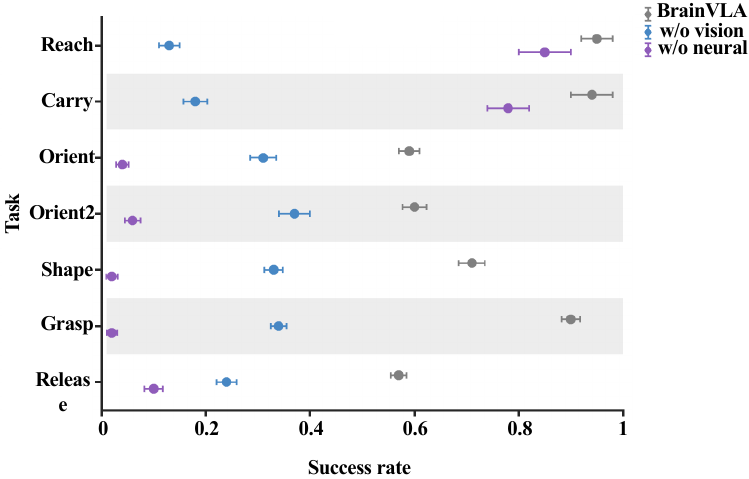}
\end{center}
\caption{Task-level analysis on D1. Task success rates under the full BrainVLA model and ablations without vision or neural input. Reach and Carry are more sensitive to vision removal, while other tasks rely more strongly on neural conditioning. Error bars denote variability across test sessions.}
\label{appen:result_per_dim}
\end{figure}

\subsection{Task-Level Analysis}

We further analyze BrainVLA on D1 at the level of individual task types, as shown in Fig.~\ref{appen:result_per_dim}. The full model achieves consistently high success rates across the evaluated tasks, indicating that BrainVLA can recover both coarse spatial motion and more fine-grained hand configurations.

In the dimension-level analysis, the modality ablations reveal a clear functional distinction between vision and neural activity as shown in \ref{fig:data_efficiency} (B). Removing vision causes the largest degradation in the translational dimensions $(x,y,z)$, which is consistent with visual observations providing direct information about spatial configuration and object-target geometry. In contrast, removing neural activity has a substantially larger effect on roll and finger-related dimensions, including upper-finger, lower-finger, and thumb control. These variables are less directly observable from the scene and depend more strongly on the intended motor command encoded in neural activity.

A similar pattern emerges at the task level. Tasks such as \emph{Reach} and \emph{Carry}, which depend strongly on spatial displacement, are particularly sensitive to the removal of vision. By comparison, orientation-, shape-, grasp-, and release-related tasks degrade more strongly when neural input is removed, reflecting their greater dependence on fine-grained hand configuration and internally specified motor intent. Overall, these results suggest that vision and neural activity provide complementary information: visual observations primarily support estimation of the external task state, whereas neural activity contributes intention-related information required for more subtle motor control.

\begin{figure}[h]
\begin{center}
\includegraphics[width=1.0\linewidth]{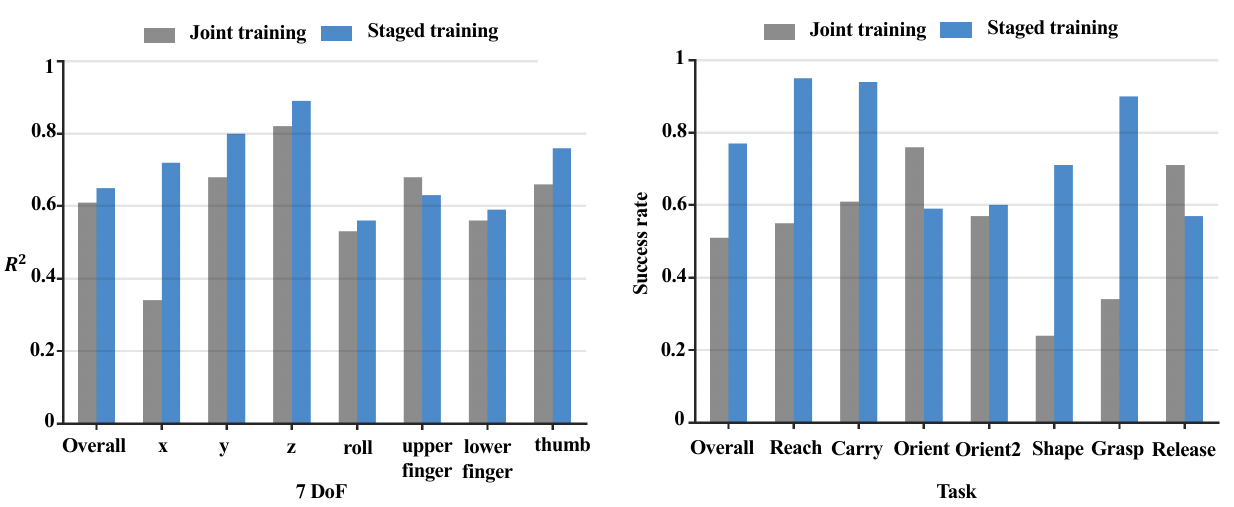}
\end{center}
\caption{Comparison of joint and staged training on D1. Left: overall and dimension-wise action $R^2$ for the 7-DoF action space.
Right: overall and task-specific success rates. Staged training provides stronger and more consistent performance overall, supporting the sequential adaptation strategy used in BrainVLA.}
\label{fig:joint_vs_staged}
\end{figure}

\subsection{Staged vs.\ Joint Training}

We further examine whether the VLA adaptation and neural conditioning stages should be optimized jointly or sequentially. In our default \emph{staged training} strategy, we first adapt the VLA policy using language conditioning as described in Fig.~\ref{fig:framework}(B), and subsequently train the neural encoder to replace language conditioning in Fig.~\ref{fig:framework}(C). In contrast, \emph{joint training} skips the language-conditioned adaptation stage: the neural encoder, LoRA adapters, action head, and proprioceptive projector are optimized simultaneously, with neural representations directly conditioning the OpenVLA backbone throughout training.

As shown in Fig.~\ref{fig:joint_vs_staged}, staged training yields stronger overall trajectory decoding and task success on D1, with particularly clear improvements for spatial dimensions and several manipulation tasks. Joint training remains competitive for some dimensions and tasks, but exhibits less consistent performance across the full action space. These results support our choice to first establish a task-adapted VLA policy and then learn the neural-to-policy interface, rather than adapting both simultaneously from neural supervision.

One possible explanation is that language provides a more stable semantic conditioning signal during VLA adaptation, allowing the policy to first acquire the target visuomotor behavior before introducing the noisier and more heterogeneous neural representation. The subsequent neural-language alignment can then map neural activity into an already established conditioning space, reducing the burden of jointly learning task adaptation and neural conditioning.

\begin{figure}[h]
\begin{center}
\includegraphics[width=1.0\linewidth]{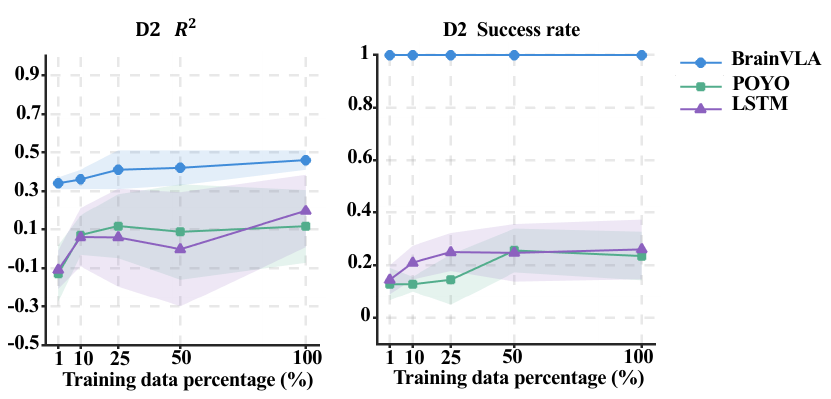}
\end{center}
\caption{$R^2$ and task success rate with scaled training D2 data. BrainVLA consistently outperforms POYO and LSTM across data regimes and remains effective under limited neural supervision. Shaded regions indicate variability across test sessions.}
\label{appen:data_efficiency_m2}
\end{figure}

\subsection{Data-efficient validation on D2}
We also evaluate data efficiency by training each model with diverse fractions of the D2 training set. As shown in Fig.~\ref{appen:data_efficiency_m2}, BrainVLA consistently outperforms POYO and LSTM across data regimes in both trajectory $R^2$ and task success rate. BrainVLA improves steadily with increasing training data, while the baselines exhibit substantially lower performance and greater variability, particularly in the low-data regime. Notably, BrainVLA retains meaningful decoding and task-level success even when trained with only a small subset of data. These results suggest that our policy improves the data efficiency of neural motor decoding.

\subsection{Neural–Language Alignment Visualization}

\begin{figure}[t]
\begin{center}
\includegraphics[width=0.8\linewidth]{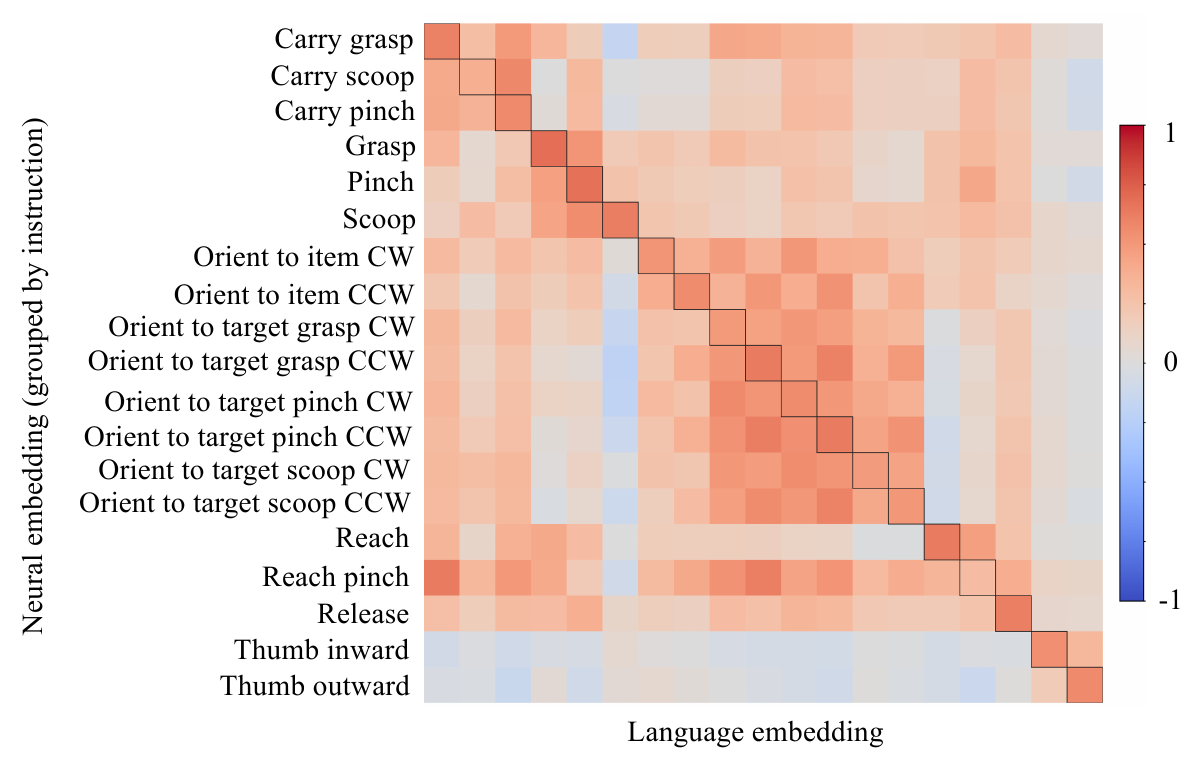}
\end{center}
\caption{Alignment between neural representation and language representation. Warmer colors denote higher representational similarity.}
\label{appen:alignment_h1}
\end{figure}

Figure~\ref{appen:alignment_h1} visualizes the alignment between neural representations and the representations of their corresponding language instructions. For each instruction, the neural latent representations are mean-pooled across latent slots, while the language token representations are mean-pooled over valid instruction tokens. Both pooled representations are $\ell_2$-normalized, and each entry in the similarity matrix is computed using their cosine similarity. Rows correspond to neural representations grouped by instruction, and columns correspond to language representations. The outlined diagonal entries indicate the matched neural-language pairs, while warmer colors denote higher representational similarity.

\end{document}